\documentclass{article}

\usepackage{microtype}
\usepackage{caption}
\usepackage{subcaption}
\usepackage{booktabs} 

\usepackage[
        hidelinks,
        colorlinks=false,
        pdftex,
        pdfpagelabels,
        bookmarks,
        hyperindex,
        hyperfigures,
        linktoc=all,
        bookmarksnumbered=true,
        bookmarksopen=true
    ]{hyperref}

\usepackage[T1]{fontenc}    
\usepackage{amsfonts}       
\usepackage{amssymb}
\usepackage{amsmath}
\usepackage{nicefrac}       

\usepackage[utf8]{inputenc}

\usepackage[pdftex]{graphicx}
\graphicspath{{images/}}
\DeclareGraphicsExtensions{.pdf,.jpeg,.png}

\usepackage{algorithm}
\usepackage{algorithmic}

\usepackage{url}

\usepackage[dvipsnames,table]{xcolor}

\usepackage{wrapfig}

\usepackage{natbib}

\usepackage{tikz}
\usetikzlibrary{bayesnet}

\usepackage{environ}
\makeatletter
\newsavebox{\measure@tikzpicture}
\NewEnviron{scaletikzpicturetowidth}[1]{%
  \def\tikz@width{#1}%
  \begin{lrbox}{\measure@tikzpicture}%
  \BODY
  \end{lrbox}%
  \pgfmathparse{#1/\wd\measure@tikzpicture}%
  \BODY
}
\makeatother

\usepackage{xr}

\makeatletter
\newcommand*{\addFileDependency}[1]{
  \typeout{(#1)}
  \@addtofilelist{#1}
  \IfFileExists{#1}{}{\typeout{No file #1.}}
}
\makeatother

\newcounter{docpart}

\newcounter{olddocpart}

\usepackage{wrapfig}
\usepackage{multirow}
\usepackage[toc,page]{appendix}

\usepackage{makecell}
\usepackage{placeins}

\usepackage{colortbl}
\definecolor{Gray}{gray}{0.9}

\usepackage{stackengine}
\usepackage{bbm}
\usepackage{arydshln}

\newcommand{\bs}{\boldsymbol}
\newcommand{\namedcomment}[3]{}

\newcommand{\eg}{{\em e.g.}}
\newcommand{\ie}{{\em i.e.}}

\newcommand{\x}{{\bs{x}}}
\renewcommand{\xi}{\x^{i}}

\newcommand{\z}{{\bs z}}

\newcommand{\params}{{\bs \theta}}

\newcommand{\M}{\mathcal{M}}

\newcommand{\Msamp}{\M_{\mathcal{S}}}

\newcommand{\pjoint}{\mathcal{P}}

\newcommand{\pdec}{p}
\newcommand{\penc}{q}

\newcommand{\Mdec}{\pdec_{\params}}
\newcommand{\Menc}{\penc_{\params}}

\newcommand{\AMIMloss}{\mathcal{L}_\text{A-MIM}}
\newcommand{\EAMIMloss}{\hat{\mathcal{L}}_\text{A-MIM}}
\newcommand{\EMIMloss}{\hat{\mathcal{L}}_\text{MIM}}

\newcommand{\E}[2]{\mathbb{E}_{#1}\left[#2\right]}

\newcommand{\HD}[2]{H_{#1} \left( \, #2 \, \right)}
\newcommand{\CE}[2]{CE \left( \, #1 \,,\, #2 \, \right)}

\newcommand{\Penc}{\Menc(\z | \x)\, \pjoint(\x)}

\usepackage[preprint]{icml2020}

\icmltitlerunning{SentenceMIM}

\begin{document}

\twocolumn[
\icmltitle{
SentenceMIM: A Latent Variable Language Model \\
}



\icmlsetsymbol{equal}{*}

\begin{icmlauthorlist}
\icmlauthor{Micha Livne}{uoft,vector}
\icmlauthor{Kevin Swersky}{goo}
\icmlauthor{David J.~ Fleet}{uoft,vector,goo}
\end{icmlauthorlist}

\icmlaffiliation{uoft}{Department of Computer Science, University of Toronto}
\icmlaffiliation{vector}{Vector Institute}
\icmlaffiliation{goo}{Google Research}

\icmlcorrespondingauthor{Micha Livne}{mlivne@cs.toronto.edu}
\icmlcorrespondingauthor{Kevin Swersky}{kswersky@google.com}
\icmlcorrespondingauthor{David J.~ Fleet}{fleet@cs.toronto.edu}

\icmlkeywords{Machine Learning, Natural Language Processing, Representation Learning, ICML}
\vskip 0.3in
]



\printAffiliationsAndNotice{}  

\begin{abstract}
SentenceMIM is a probabilistic auto-encoder for language data, 
trained with Mutual Information Machine (MIM) learning 
to provide a fixed length representation of variable length 
language observations (\ie, similar to VAE).
Previous attempts to learn VAEs for language data faced challenges 
due to posterior collapse. MIM learning encourages high mutual 
information between observations and latent variables, 
and is robust against posterior collapse.  As such, it learns 
informative representations whose dimension can be an order of 
magnitude higher than existing language VAEs.
Importantly, the SentenceMIM loss has no hyper-parameters, simplifying optimization.
We compare sentenceMIM with VAE, and AE on multiple datasets. 
SentenceMIM yields excellent reconstruction, comparable to AEs,
with a rich structured latent space, comparable to VAEs. 
The structured latent representation is demonstrated with interpolation 
between sentences of different lengths.
We demonstrate the versatility of sentenceMIM by utilizing a trained model for
question-answering and transfer learning, without fine-tuning, outperforming 
VAE and AE with similar architectures.
\end{abstract}

\section{Introduction}

Generative modelling of text has become one of the predominant approaches to natural language processing (NLP), particularly in the machine learning community. 
It is favoured because it supports probabilistic reasoning and it provides a principled framework for unsupervised learning in the form of maximum likelihood. 
Unlike computer vision, where various generative approaches have proliferated~\cite{dinh2016density,goodfellow2014generative,Kingma2013,oord2016pixel,Rezende2014,vahdat2020NVAE}, 
current methods for text mainly rely on auto-regressive models (\eg, \citet{brown2020language}).

Generative latent variable models (LVMs), such as the variational auto-encoder 
(VAE) \cite{Kingma2013,Rezende2014}, are versatile and have been 
successfully applied to myriad domains.  
Such models consist of an encoder, which maps observations to distributions over 
latent codes, and a decoder that maps latent codes to distributions over observations.
LVMs are widely used and studied because they can learn a latent 
representation that carries many useful properties. 
Observations are encoded as fixed-length vectors that capture salient information, 
allowing for semantic comparison, interpolation, and search.
They are often useful in support of downstream tasks, such as transfer or k-shot learning. 
They are also often interpretable, capturing distinct factors of variation in different latent dimensions.
These properties have made LVMs especially compelling in the vision community.

Despite their desirable qualities, generative LVMs have not 
enjoyed the same level of success with text data. 
There have been recent proposals to adapt VAEs to text~\cite{DBLP:journals/corr/BowmanVVDJB15,  Guu2017GeneratingSB, KruengkraiACL2019,LiEtAlICNLG2019, DBLP:journals/corr/YangHSB17,bosc-vincent-2020-sequence}, 
but despite encouraging progress, they have not reached the same level of performance on natural 
language benchmarks as auto-regressive models
(\eg, \citep{DBLP:journals/corr/abs-1708-02182,DBLP:journals/corr/abs-1803-10049,pmlr-v97-wang19f}).
This is often attributed to the phenomenon of posterior collapse \citep{Fang_iLVM_2019_EMNLP,Li2019ASV},
in which the decoder captures all of the modelling power and the encoder conveys little to no information. 
For text, where the decoder is naturally auto-regressive, this has proven challenging to mitigate. 
A notable exception by \citet{li2020_Optimus} utilizes pre-trained BERT encoder
\citep{devlin-etal-2019-bert} and GPT-2 decoder \citep{Radford2019LanguageMA} in order to train a powerful VAE model. 
While showing strong PPL results, the training requires carefully designed heuristics to reduce posterior collapse.

This paper introduces sentenceMIM (sMIM), a new LVM for text. We use the Mutual Information Machine (MIM) framework 
by \citet{2019arXiv191003175L} for learning, and base our architecture on \citet{DBLP:journals/corr/BowmanVVDJB15}.
MIM is a recently introduced LVM framework that shares the same underlying architecture 
as VAEs, but uses a different learning objective that is robust against posterior collapse. 
MIM learns a highly informative and compressed latent representation, and often strictly 
benefits from more powerful architectures. 

We argue that an ideal LVM should be able to capture all aspects of variation of variable-size text observations 
within a fixed-size latent representation.
As such, high-dimensional latent codes are required, which is challenging with VAEs.
An ideal  model should provide excellent reconstruction, with fixed-size codes for variable-length sentences,
and be useful for various downstream tasks.

Auto-encoders (AEs)  \cite{NIPS1993_9e3cfc48} provide excellent reconstruction, 
but lack useful semantic structure in the learned representations.
AEs also allow one to learn high dimensional latent codes, only limited in practice by over-fitting.
VAEs, on the other hand, encourage semantic representations by regularizing the distribution over latent codes to match a given prior.
Such regularization, however, also contributes to posterior collapse, limiting the dimension of latent codes and reconstruction quality.
Here we propose MIM learning to enable high dimensional representations, while encouraging low latent entropy (under certain conditions) to promote clustering of semantically similar observations.
By encouraging the latent codes to match a known distribution we preserve the ability generate samples.
The resulting model offers a learned representation with high mutual information (\ie, to capture aspects of 
variation in the data), with low marginal entropy (\ie, introducing semantic structure to the learned representation), 
while aligning the latent distribution with a known prior.
sMIM also requires no hyper-parameter tuning for the loss, similar to AEs, which simplifies training.

This paper explores and contrasts properties of VAE learning, MIM learning, and AE learning on four well-known
text datasets, all with similar architectures.
We show that sMIM provides better reconstruction than VAE models, matching the reconstruction accuracy of AEs,
but with semantically meaningful representations, comparable to VAEs.
We further demonstrate the quality of the sMIM representation by generating diverse 
samples around a given sentence and interpolating between sentences. 
Finally, we show the versatility of the learned representation by applying a pre-trained sMIM 
model to a question answering task with state-of-art performance as compared to single task, supervised models.


\section{Problem Formulation}

Let $\x \in \mathcal{X} = \{\x_i\}_{i=1}^{X}$ be a discrete variable representing
a sequence of $T$ tokens from a finite vocabulary $\mathcal{V}$.
A sequence might be a sentence or a paragraph, for example.
The set $\mathcal{X}$ comprises all sequences we aim to model.
The size of $\mathcal{X}$, \ie, $X$, is typically unknown and large.
Let $\pjoint(\x)$ be the unknown probability of sentence $\x \in \mathcal{X}$.

We model the distribution over a sequence of length $T$ as an auto-regressive distribution over $T+1$ tokens, 
where the additional end-of-text special token\footnote{\textsc{<BOT>}, \textsc{<EOT>} are a special beginning/end-of-text tokens. The token \textsc{<UNK>} represents an out-of-vocabulary word.},  \textsc{<EOT>}, 
effectively captures the probability that the sequence length is $T$.
More explicitly, we model the following distribution
\begin{equation}
    \pjoint(\x) = \prod_{k=0}^{T} \pdec(x^k | \x^{<k})
\end{equation}
where $\x^{<k}$ denotes the tokens preceding $x^k$,
$\pdec(x^k | \x^{<k})$ is a categorical distribution over $\mathcal{V}$, 
and $\pdec(x^{T} = \text{\textsc{<EOT>}} | \x^{<T})$ is the probability that $T$ is the sentence length.

We learn a latent variable model given $N$ fair samples from $\pjoint(\x)$, where $N \ll X$, with
discrete observations $\x \in \mathcal{X}$, and a continuous latent space $\z \in \mathbb{R}^{d}$.
The encoder, $\Menc(\z | \x)$, maps sequences to a distribution over continuous latent codes,
and a corresponding decoder, $\Mdec(\x|\z)$, maps a latent code to a distribution over sequences.
Let $\params$ be the joint parameters of the encoder and decoder.




\subsection{Encoder-Decoder Specification} \label{sec:nlp-formulation}

\begin{figure}[t]
    \centering
    \scalebox{0.85}{\begin{tikzpicture}


\node[latent]         (z) {$\z$};

\node[latent, right=of z]                          (xdec1) {$h^1_d$};
\node[const, above=of xdec1, yshift=-0.5cm]        (xdec1_token) {\small $\Mdec^{seq}(x^1|h^1_d)$};
\node[const, above=of xdec1, yshift=0.25cm]         (xdec1_token_) {the};
\node[latent, right=of xdec1]          (xdec2) {$h^2_d$};
\node[const, above=of xdec2, yshift=-0.5cm]        (xdec2_token) {\small $\Mdec^{seq}(x^2|h^2_d)$};
\node[const, above=of xdec2, yshift=0.25cm]        (xdec2_token_) {cat};
\node[latent, right=of xdec2]          (xdec3) {$h^3_d$};
\node[const, above=of xdec3, yshift=-0.5cm]        (xdec3_token) {\small $\Mdec^{seq}(x^3|h^3_d)$};
\node[const, above=of xdec3, yshift=0.25cm]        (xdec3_token_) {is};
\node[latent, right=of xdec3]          (xdec4) {$h^4_d$};
\node[const, above=of xdec4, yshift=-0.5cm]        (xdec4_token) {\small $\Mdec^{seq}(x^4|h^4_d)$};
\node[const, above=of xdec4, yshift=0.18cm]        (xdec4_token_) {sitting};
\node[latent, right=of xdec4]          (xdec5) {$h^5_d$};
\node[const, above=of xdec5, yshift=-0.5cm]        (xdec5_token) {\small $\Mdec^{seq}(x^5|h^5_d)$};
\node[const, above=of xdec5, yshift=0.18cm]        (xdec5_token_) {<EOT>};

\node[const, below=of xdec1, yshift=-0.5cm]        (xenc1_token) {<BOT>};
\node[const, below=of xdec2, yshift=-0.5cm]        (xenc2_token) {the};
\node[const, below=of xdec3, yshift=-0.5cm]        (xenc3_token) {cat};
\node[const, below=of xdec4, yshift=-0.5cm]        (xenc4_token) {is};
\node[const, below=of xdec5, yshift=-0.5cm]        (xenc5_token) {sitting};

\edge {z} {xdec1}
\edge[bend right=35] {z} {xdec2}
\edge[bend right=35] {z} {xdec3}
\edge[bend right=35] {z} {xdec4}
\edge[bend right=35] {z} {xdec5}

\draw[->, >={triangle 45}] (xdec1) -- (xdec2) node [label=above left:{GRU ~~~~}] {} ;
\draw[->, >={triangle 45}] (xdec2) -- (xdec3) node [label=above left:{GRU ~~~~}] {} ;
\draw[->, >={triangle 45}] (xdec3) -- (xdec4) node [label=above left:{GRU ~~~~}] {} ;
\draw[->, >={triangle 45}] (xdec4) -- (xdec5) node [label=above left:{GRU ~~~~}] {} ;

\edge {xdec1} {xdec1_token}
\edge {xdec2} {xdec2_token}
\edge {xdec3} {xdec3_token}
\edge {xdec4} {xdec4_token}
\edge {xdec5} {xdec5_token}

\edge {xenc1_token} {xdec1}
\edge {xenc2_token} {xdec2}
\edge {xenc3_token} {xdec3}
\edge {xenc4_token} {xdec4}
\edge {xenc5_token} {xdec5}

\end{tikzpicture}}
    \vspace*{-0.6cm}
    \caption{The decoder, implemented with GRU, is auto-regressive and conditioned on latent code $\z$.
    Words are represented by parametric embeddings.
    At each step the inputs are the latent code and previous output token.
    The GRU output provides a categorical distribution over tokens $\x^k$,    
    from which the next token is sampled.
    }
    \label{fig:sentence-mim-arch-decoder}
\end{figure}
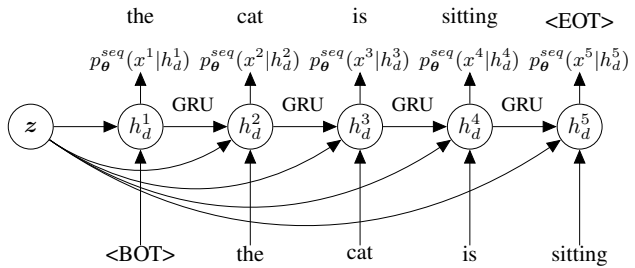

In what follows we adapt the architecture proposed by \citet{DBLP:journals/corr/BowmanVVDJB15}, 
the main difference being the use of GRUs \citep{cho-etal-2014-learning} instead of LSTMs \citep{Hochreiter:1997:LSM:1246443.1246450}.
We opt for a simple architecture instead of more recent variants, such as Transformers \citep{Vaswani2017}
or AWD-LSTMs \citep{merity2018regularizing}, to focus on the effect of the learning framework on a given architecture (\ie, MIM, VAE, AE), 
rather than the architecture itself.

Beginning with the generative process, let $\Mdec(\x|\z) $ be a conditional auto-regressive distribution 
over a sequence of $T$ tokens, $\x = (x^{0},\ldots, x^{T-1}, x^{T} = \text{\textsc{<EOT>}})$,
\begin{equation}
    \log \Mdec(\x|\z) \, =\, \sum_{k=0}^{T} \log \Mdec(x^k\, |\, \x^{<k}, \z)
    \label{eq:p-x-seq-given-z}
\end{equation}
where $\Mdec(x^k | \cdot )$ is a categorical distribution over $|\mathcal{V}|$
possible tokens for the $k^{th}$ element in $\x$.
According to the model (Fig.\ \ref{fig:sentence-mim-arch-decoder}), generating
a sentence $\x$ with latent code $\z$ entails sampling each token from a
distribution conditioned on the latent code and previously sampled tokens.
Tokens are modelled with a parametric embedding. Conditioning the distribution over $\z$
entails concatenating $\z$ to the input embeddings per token \citep{DBLP:journals/corr/BowmanVVDJB15}.

\begin{figure}[t]
    \vspace*{-0.01cm}
    \centering
    \scalebox{0.85}{\begin{tikzpicture}


\node[latent]                          (xenc1) {$h^1_e$};
\node[const, below=of xenc1, yshift=0.5cm]        (xenc1_token) {<BOT>};
\node[latent, right=of xenc1]          (xenc2) {$h^2_e$};
\node[const, below=of xenc2, yshift=0.5cm]        (xenc2_token) {the};
\node[latent, right=of xenc2]          (xenc3) {$h^3_e$};
\node[const, below=of xenc3, yshift=0.5cm]        (xenc3_token) {cat};
\node[latent, right=of xenc3]          (xenc4) {$h^4_e$};
\node[const, below=of xenc4, yshift=0.5cm]        (xenc4_token) {is};
\node[latent, right=of xenc4]          (xenc5) {$h^5_e$};
\node[const, below=of xenc5, yshift=0.5cm]        (xenc5_token) {sitting};

\node[const, right=of xenc5]         (z) {$\Menc(\z|h^5_e)$};


\draw[->, >={triangle 45}] (xenc1) -- (xenc2) node [label=above left:{GRU ~~~}] {} ;
\draw[->, >={triangle 45}] (xenc2) -- (xenc3) node [label=above left:{GRU ~~~}] {} ;
\draw[->, >={triangle 45}] (xenc3) -- (xenc4) node [label=above left:{GRU ~~~}] {} ;
\draw[->, >={triangle 45}] (xenc4) -- (xenc5) node [label=above left:{GRU ~~~}] {} ;

\edge {xenc1_token} {xenc1}
\edge {xenc2_token} {xenc2}
\edge {xenc3_token} {xenc3}
\edge {xenc4_token} {xenc4}
\edge {xenc5_token} {xenc5}

\edge {xenc5} {z}

\end{tikzpicture}}
     \vspace*{-0.6cm}
    \caption{The encoder is implemented with GRU.
    Each word is represented by a parametric embedding.
    Given the input sequence, the encoder maps the last hidden state 
    to the mean and variance of a Gaussian posterior over latent codes.
    }
    \label{fig:sentence-mim-arch-encoder}
    \vspace*{-0.1cm}
\end{figure}
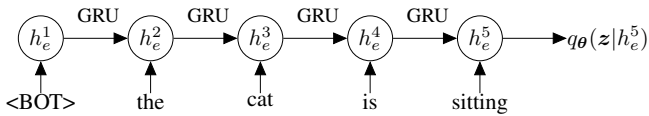

The encoder $\Menc(\z|\x)$ is the posterior distribution over the latent variable $\z$,
conditioned on a sequence $\x$.  We take this to be Gaussian whose mean and diagonal 
covariance are specified by mappings $\mu_\params$ and  $\sigma_\params$:
\begin{equation}
    \Menc(\z|\x) \, =\, \mathcal{N}(\z; \mu_\params(\x), \sigma_\params(\x))
    \label{eq:nlp-q-z-given-x}
\end{equation}
Linear mappings $\mu_\params$ and  $\sigma_\params$ are computed from the last hidden state of a GRU (see Fig.\ \ref{fig:sentence-mim-arch-encoder}).


\subsection{MIM Learning Objective}
\label{sec:nlp-learning}

The Mutual Information Machine (MIM) \cite{2019arXiv191003175L} is a versatile generative LVM 
which can be used for representation learning, and sample generation.
MIM learns a model with high mutual information between observations and latent codes,
and is robust against posterior collapse. 

The MIM framework begins with two {\em anchor} distributions, $\pjoint(x)$ and $\pjoint(z)$, for
observations and the latent space, from which one can draw samples.  They are fixed and not learned.
MIM also has a parameterized encoder-decoder pair, $\Menc(\z|\x)$ and $\Mdec(\x|\z)$,
and parametric marginal distributions $\Menc(\x)$ and $\Mdec(\z)$.
These parametric elements define joint encoding and decoding {\em model} distributions:
\begin{eqnarray}
    \Menc(\x,\z) & = & \Menc(\z|\x)~ \Menc(\x) ~, \\
    \Mdec(\x,\z) & = & \Mdec(\x|\z)~\Mdec(\z) ~ .
\end{eqnarray}
MIM learning entails the minimization of the cross-entropy between a sample distribution 
and the model encoding and decoding distributions \citep{2019arXiv191003175L}. 
This simple loss constitutes a variational upper bound on a regularized Jensen-Shannon divergence, 
resembling VAE in which a model distribution matches samples from a sample distribution via KL 
divergence minimization \citep{Zhao_Song_Ermon_2018}. 
Fundamentally, MIM learning differs from VAE learning, with the former being an upper bound on the 
joint cross-entropy, while the latter being an upper bound on the marginal cross-entropy of the observations.

MIM requires sampling from the decoder during training, which can be slow for sequential computational models.
For language modeling we therefore use A-MIM learning, a MIM variant that minimizes a loss
defined on the encoding and decoding distributions, with samples drawn from an encoding 
{\em sample} distribution, denoted  $\Msamp^{\penc}(\x, \z)$;  \ie,
\begin{eqnarray}
    \Msamp^{\penc}(\x, \z) ~=~ \Penc \label{eq:sample-dist} ~.
\end{eqnarray}
The A-MIM loss is defined as follows,
\begin{eqnarray}
    \AMIMloss(\params) &=&
    \frac{1}{2} \big(\, \CE{\Msamp^{\penc}(\x, \z)}{\Menc \left(\x, \z \right)} \label{eq:A-MIM-asymmetric-bound} \\
    && \quad + ~ \CE{\Msamp^{\penc}(\x, \z)}{\Mdec \left(\x, \z \right)} \, \big) \nonumber \\
    &\geq& H_{\Msamp^{\penc}} (\x) + H_{\Msamp^{\penc}} (\z)  - I_{\Msamp^{\penc}} (\x;\z)~,
    \nonumber
\end{eqnarray}
where $\CE{\cdot}{\cdot}$ is cross-entropy, $H_{\Msamp^{\penc}}(\cdot)$ is information entropy over
distribution $\Msamp^{\penc}$, and $I(\cdot;\cdot)$ is mutual information.
Minimizing $\AMIMloss(\params)$ learns a model with a consistent
encoder-decoder, high mutual information, and low marginal entropy \citep{2019arXiv191003175L}.
The A-MIM loss is in fact a variational upper bound on the joint entropy of the encoding 
sample distribution.


\subsection{Implicit and Explicit Model Marginals} \label{sec:nlp-variational-priors}

To complete the model specification, we define the model marginals $\Menc(\x)$ and $\Mdec(\z)$.
We call the marginal {\it explicit}  when we can evaluate the probability of a sample 
under the corresponding distribution, and {\it implicit} otherwise.

Examples of explicit marginals are a Gaussian for $\Mdec(\z)$, and an auto-regressive distribution for $\Menc(\x)$. 
They enable evaluation of the probability of a sample straightforwardly. 
However, the inductive bias in such distributions, or the architecture, can lead to a  challenging optimization problem. 

An implicit marginal distribution can be defined via marginalization of a joint distribution.  
To help encourage consistency, and avoid introducing more model parameters, one can
define model marginals in terms of the sample distributions, like $\Msamp^{\penc}(\x, \z)$ above \citep{DBLP:journals/corr/BornscheinSFB15,2019arXiv191003175L,DBLP:journals/corr/TomczakW17}.
They allow one to share parameters between a marginal and the corresponding conditional distribution, 
and to reduce the inductive bias in the architecture. 
Unfortunately, evaluating the probability of a sample under an implicit distribution is intractable in general.

We define $\Menc(\x)$ as a marginal over the decoder \cite{DBLP:journals/corr/BornscheinSFB15}; \ie,
\begin{equation}
     \Menc(\x) \, = \, \E{\pjoint(\z)}{\Mdec(\x|\z)}  ~,
    \label{eq:nlp-q-x}
\end{equation}
where the latent anchor is defined to be a standard normal, $\pjoint(\z) = \mathcal{N}(\z ; 0, 1)$.
Similarly, the model density over latent codes is defined as
\begin{equation}
    \Mdec(\z) \,=\,  \E{\pjoint(\x)}{\Menc(\z|\x)}~.
    \label{eq:nlp-p-z}
\end{equation}
\ie, the latent marginal is defined as the aggregated posterior, in the spirit of the 
VampPrior \cite{DBLP:journals/corr/TomczakW17} and Exemplar VAE \cite{ExemplarVAE2020}.

\subsection{Tractable Bounds to Loss}

Given training data $D = \{\x_i \}_{i=1}^N$, the empirical loss is
\begin{eqnarray}
    \!\EAMIMloss (\params) \!\!\!\!
    &=& \!\!\!
    - \frac{1}{2N}
    \sum_{\x_i}\! \E{\Menc(\z|\x_i)}{\log  \Menc(\z|\x_i)\,\Menc(\x_i)} \nonumber \\
    && \hspace*{-0.65cm}
    - ~ \frac{1}{2N}
    \sum_{\x_i}\! \E{\Menc(\z|\x_i)}{\log \Mdec(\x_i|\z)\,\Mdec(\z)} ~
    \label{eq:nlp-mim-loss-empirical}
\end{eqnarray}
where $\sum_{\x_i}$ denotes a sum over $N$ fair samples drawn from $\pjoint(\x)$, 
as a MC approximation to expectation over $\pjoint(\x)$.

Unfortunately, the empirical loss in Eqn.\ \eqref{eq:nlp-mim-loss-empirical} is intractable since we
cannot evaluate the log-probability of the marginals $\Mdec(\z)$ and $\Menc(\x)$.
In what follows we obtain a tractable empirical bound on the loss in Eqn.\ \eqref{eq:nlp-mim-loss-empirical}
for which, with one joint sample, we obtain an unbiased and low-variance estimate of the gradient using reparameterization \cite{Kingma2013}.

We first derive a tractable lower bound to $\log \Menc(\x_i)\, $:
\begin{eqnarray}
    \log \Menc(\x_i)\!\! &\underset{\text{\tiny (Eqn. \ref{eq:nlp-q-x})}}{=}& \!\log \E{\pjoint(\z)}{\Mdec(\x_i|\z)}  \label{eq:nlp-q-x-lower-bound} \\
    &\underset{\text{\tiny (IS)}}{=}& \!\log \E{\Menc(\z|\x_i)}{\Mdec(\x_i|\z)\frac{\pjoint(\z)}{\Menc(\z|\x_i)}}
     \nonumber \\
    &\underset{\text{\tiny (JI)}}{\ge}& \!\E{\Menc(\z|\x_i)}{\log \left(  \Mdec(\x_i|\z)\frac{\pjoint(\z)}{\Menc(\z|\x_i)} \right) }  \nonumber
\end{eqnarray}
where the second and third lines are obtained using importance sampling and Jensen's inequality.
We remind the reader that $\Menc(\x_i)$ is a variational marginal that can depend
on $\x_i$.  Indeed, Eqn.\ \eqref{eq:nlp-q-x-lower-bound} is the usual ELBO.

To derive a lower bound to $\log \Mdec(\z)$, we begin with the following inequality,
\begin{eqnarray}
\!\!\log \E{\pjoint(\x)}{h(\x;\cdot)}  \!\!
&=& \!\!  \log \sum_{i} \pjoint(\x_i)\, h(\x_i;\cdot) \nonumber  \\
&\ge& \!\log \pjoint(\x')\, h(\x';\cdot) ~,
\label{eq:log-lower-bound}
\end{eqnarray}
for any sample $\x'$, any discrete distribution $\pjoint(\x)$, and any non-negative function $h(\x;\cdot) \ge 0$.
The inequality in Eqn.\ \eqref{eq:log-lower-bound} follows from $\log a \ge \log b$ for $a \ge b$.
Using this bound, we express a lower bound to $\Mdec(\z)$ as follows,
\begin{eqnarray}
    \log \Mdec(\z) &\underset{\text{\tiny (Eqn. \ref{eq:nlp-p-z})}}{=}& \log \E{\pjoint(\x)}{\Menc(\z|\x)} \nonumber \\
    &\underset{\text{\tiny (Eqn. \ref{eq:log-lower-bound})}}{\ge}& \log \Menc(\z|\x')  + \log \pjoint(\x') \label{eq:nlp-p-z-lower-bound}
\end{eqnarray}
for any sample $\x'$. During training, given a joint sample $~\x_i, \z_i \sim \Penc$, we choose $\x' = \x_i$.

\begin{figure}[t]
\vspace*{-0.35cm}
\centering
\begin{minipage}[t]{0.95\columnwidth}
\begin{algorithm}[H]
    \caption{Learning parameters $\params$ of sentenceMIM}
    \label{algo:sentencemim}
    \begin{algorithmic}[1]
        \WHILE{not converged}
        \STATE $D_\mathrm{enc} \gets \{ \x_j, \z_j \sim \Menc(\z|\x)\pjoint(\x) \}_{j=1}^{N}$
        \STATE $\EMIMloss \left( \params ; D \right) = -\frac{1}{N}\! \sum_{i=1}^{N}\! \big( ~ \log \Mdec(\x_i | \z_i) ~~~~~~
        \linebreak ~~~~~~~~~~~~~~~~~~  + \frac{1}{2} \left( \log \Menc(\z_i | \x_i) + \log \pjoint(\z_i) \right) ~ \big)$
        \STATE $\Delta \params \propto -\nabla_{\params}  \EMIMloss \left( \params ; D \right)$
        \COMMENT{\textcolor{gray}{\textit{Gradient computed through sampling using reparameterization}}}
        \ENDWHILE
    \end{algorithmic}
\end{algorithm}
\end{minipage}
\vspace*{-0.5cm}
\end{figure}

Substituting Eqns.\ \eqref{eq:nlp-q-x-lower-bound} and \eqref{eq:nlp-p-z-lower-bound}
into Eqn.\ \eqref{eq:nlp-mim-loss-empirical} gives the final form of an
upper bound on the empirical loss; \ie,
\begin{eqnarray}
    \EAMIMloss \!\!
    & \le & \!\! -\frac{1}{N} \, \sum_{i} \E{\Menc(\z |\x_i)}{\log \Mdec(\x_i|\z)}  \nonumber \\
    & & -\frac{1}{2N} \, \sum_{i} \E{\Menc(\z |\x_i)}{\log \big( ~ \Menc(\z|\x_i) \pjoint(\z) ~ \big) } \nonumber \\
    & & + \frac{1}{2}\HD{\pjoint}{\x} ~. \label{eq:nlp-mim-loss-empirical-bound}
\end{eqnarray}
We find an unbiased, low variance estimate of the gradient of Eqn.\ \eqref{eq:nlp-mim-loss-empirical-bound}
with a single joint sample $\z_i, \x_i \sim \Penc$ and reparameterization.
The last term, $\HD{\pjoint}{\x}$, is a constant, independent of model parameters
and can therefore be ignored during optimization.
The resulting learning process is described in Algorithm\ \ref{algo:sentencemim}.

To better understand the proposed bounds, we note that A-MIM achieves good
reconstruction by learning posteriors with relatively small variances
(\ie, relative to the distance between latent means).
Our choice of $\x'=\x_i$ exploits this, allowing good gradient estimation,
and facilitating fast convergence.
We further provide empirical evidence for these properties below in Fig.\ \ref{fig:nlp-dist-ptb}.

\section{Experiments} \label{sec:nlp-experiments}

\subsection{Datasets} \label{sec:nlp-datasets}

\begin{table}[th]
    \centering
    \setlength{\tabcolsep}{0.2em} 
    {\small
    \renewcommand{\arraystretch}{1.2}
\begin{tabular}{l||ccccc}
 \hline
 & \multicolumn{3}{c}{Sentences} &  &  \\
 \hline
 (word level) & Train & Valid. & Test & Vocab. & \#words (avg.)  \\
 \hline \hline
PTB  & 42068 & 3370 & 3761 & 9877 & 21 $\pm$ 10 \\
Yahoo & 100K & 10K & 10K & 37165 & 76 $\pm$ 55  \\
Yelp15 & 100K & 10K & 10K & 19730 & 100 $\pm$ 51  \\
WikiText-103 & 200K & 10K & 2185 & 89247 & 115 $\pm$ 60\\
\hdashline[1pt/1pt]
Everything \textsuperscript{\textdagger} & 442067 & 33369 & 33760 & 105965 & 94 $\pm$ 60 \\
\hline
\end{tabular}
    }
    \vspace*{0.01cm}
    \caption{
    Dataset properties summary for Penn Tree Bank, 
    Yahoo Answers and Yelp15, 
    and sampled WikiText-103. 
    Everything \textsuperscript{\textdagger} is the union of all four datasets.
    }
    \label{tab:posterior-collapse-text-datasets}
\end{table}

We show experimental results on four word level datasets
described in Table \ref{tab:posterior-collapse-text-datasets}, namely,
Penn Tree Bank \cite{Marcus:1993:BLA:972470.972475},
Yahoo Answers and Yelp15 (following \citet{DBLP:journals/corr/YangHSB17}),
and WikiText-103 \cite{DBLP:journals/corr/MerityXBS16}.
We use  the Yahoo and Yelp15 datasets of \citet{DBLP:journals/corr/YangHSB17}, 
which draw 100k samples for training, and 10k for validation and testing.
For WT103 we draw 200k samples for training, 10k for validation, and retain the original test data. 
Empty lines and headers were filtered from the WT103 data.

\subsection{Architecture and Optimization} \label{sec:nlp-architecture}

Our auto-encoder architecture (Figs.\ \ref{fig:sentence-mim-arch-decoder} and \ref{fig:sentence-mim-arch-encoder}) 
followed that proposed by \citet{DBLP:journals/corr/BowmanVVDJB15}.
As is common, we concatenated $\z$ with the input to the decoder 
(\ie, a "context", similar to \citet{he2018lagging,DBLP:journals/corr/YangHSB17,DBLP:journals/corr/BowmanVVDJB15}).
We use the same architecture, parameterization, and latent dimensions for both sMIM and a VAE 
variant called sVAE, for comparison. 
We also trained deterministic auto-encoders with the same architecture, called sAE,
by replacing the sampled latent code with the deterministic mean of the posterior (\ie, $\z_i = \E{\z'}{\Menc(\z'|\x_i)}$).
Effectively, the only difference between these variants is the choice of loss function. Training times for all models are similar.

For PTB we trained models with 1 layer GRU, latent space dimensions of 16D, 128D, and 512D, 
a 512D hidden state, 300D word embeddings, and 50\% embedding dropout.
We trained all models with Adam \citep{Kingma2014} with initial learning rate $lr = 10^{-3}$. 
Training took less than 30 minutes on a single TITAN Xp 12G GPU.
For Yahoo Answers, Yelp15, and WT103 we trained models with 1 layer GRU, 
latent space dimensions of 32D, 512D, 1024D, a 1024D hidden state, 512D word embeddings, 
and 50\% embedding dropout.
We trained these models with SGD \citep{pmlr-v28-sutskever13}, with initial $lr = 5.0$,
and 0.25 $L_2$ gradient clipping.
All model and optimization hyper-parameters were taken from publicly available implementation of the method proposed by \citet{DBLP:journals/corr/BowmanVVDJB15}.

In all cases we use a learning rate scheduler that scaled the learning rate by 0.25
following two/one epochs (PTB/other datasets, respectively) with no improvement in
the validation loss.
We used a mini-batch size of 20 in all cases.
Following \cite{Sutskever:2014:SSL:2969033.2969173} we feed the input in reverse 
to the encoder, such that the last hidden state in the encoder depends on the 
first word of the sentence. 

We trained sVAEs with the regular ELBO, and with KL divergence annealing 
(denoted "+ kl"), where a scalar weight on the KL divergence term 
is increased from 0 to 1 over 10k mini-batches to lower the risk of posterior 
collapse
\citep{DBLP:journals/corr/BowmanVVDJB15}.
We use no loss manipulation heuristics in the optimization of sMIM or sAE.






\subsection{Information Content in the Latent Code} \label{nlp-quantitative-results}

\begin{table}[t]
    \centering
    \scalebox{1.1}{
    \setlength{\tabcolsep}{0.2em} 
    {\scriptsize
    \renewcommand{\arraystretch}{1.2}
        \begin{tabular}{l||c|c|c|c|c}
        \hline
        $\z$ dim. & Enc. Recon. $\downarrow$ & KL & Rand. Recon. & BLEU $\uparrow$ & $|\params|$ \\ \hline \hline
        sVAE (16) & 105.24 (0.12) & 1.6 & 105.91 (0.01) & 0.124 & 11M  \\
        sVAE (128) & 106.72  (0.12) & 0.64 & 106.89 (0.01)  & 0.118 & 11M  \\
        sVAE (512) & 108.52  (0.23) & 0.41  & 108.88  (0.01) & 0.116 & 12M \\
        \hline
        sAE (16) & 91.86  &  & 163.9 (0.02) & 0.348 & 11M  \\
        sAE (128) & 67.56 &  & 113.61 (0.01)  & 0.589 & 11M  \\
        sAE (512) & 62.08 &   & 102.93 (0.01) & 0.673 & 12M \\
        \hline
        sMIM (16) & 90.12 (0.03) &  & 161.037 (0.02)  & 0.35 & 11M \\
        sMIM (128) & 67.35 (0.008) & & 136.2 (0.04) & 0.61 & 11M \\
        sMIM (512) & \textbf{59.23} (0.01) & & 133.74 (0.01) & \textbf{0.679} & 12M \\
        \hdashline[1pt/1pt]
        sMIM (1024) \textsuperscript{\textdagger} & \textbf{26.43} (0.0) & &  & \textbf{0.724} & 179M \\
        \hline
        \end{tabular}
        }
    }
      \vspace*{0.1cm}
    \caption{
    Reconstruction results for {\bf PTB} are averaged over 10 runs (stdev).
    Models\textsuperscript{\textdagger} use extra training data.
    Reconstruction with a sample $\z \sim \Menc(\z|\x)$ (Enc. Recon.) and with 
    a random sample $\z \sim \mathcal{N}(\z)$ (Rand. Recon.).
    An uninformative latent space will result in similar reconstruction values.
    see text for details.
    }
    \label{tab:language-modelling-quantitative-ptb}
    \vspace*{-0.1cm}
\end{table}

\begin{table}[t]
    \centering
    \scalebox{1.1}{
    \setlength{\tabcolsep}{0.2em} 
    {\scriptsize
    \renewcommand{\arraystretch}{1.2}
        \begin{tabular}{l||c|c|c|c|c}
        \hline
        $\z$ dim. & Enc. Recon. $\downarrow$ & KL & Rand. Recon. & BLEU $\uparrow$ & $|\params|$ \\ \hline \hline
        sVAE (32) + kl & 401.63 (0.01) & 31.86  & 425.92 (0.01)  & 0.274 & 40M \\
        sVAE (512) + kl & 379.93 (0.01) & 4.19 & 385.76 (0.01) & 0.18 & 43M  \\
        sVAE (1024) + kl & 384.85 (0.01) & 3.01  & 387.63 (0.01)  & 0.176 & 46M \\
        \hline
        sAE (32) & 330.25  &  & 697.316 (0.0) & 0.388 & 40M  \\
        sAE (512) & 228.34 &  & 515.75 (0.0)  & 0.669 & 43M  \\
        sAE (1024) & 222.7  &   & 503.87 (0.0) & \textbf{0.684} & 46M \\
        \hline
        sMIM (32) & 396.34 (0.0) &  & 427.6 (0.0) & 0.309 & 40M  \\
        sMIM (512) & 220.03 (0.0) &  & 600.29 (0.0) & 0.673 & 43M \\
        sMIM (1024) & \textbf{219.37} (0.0) &  & 543.36 (0.0) & 0.676 & 46M \\
        \hdashline[1pt/1pt]
        sMIM (1024)\textsuperscript{\textdagger} & \textbf{199.72} (0.0) & & & \textbf{0.686} & 179M \\
        \hline
        \end{tabular}
        }
    }
      \vspace*{-0.1cm}
    \caption{
    Reconstruction results for {\bf Yelp15} are averaged over 10 runs.
    Models\textsuperscript{\textdagger} use extra training data
    (See Table \ref{tab:language-modelling-quantitative-yelp} for details).
    }
    \label{tab:language-modelling-quantitative-yelp}
    \vspace*{-0.1cm}
\end{table}

\begin{table}[t]
    \centering
    \scalebox{1.1}{
    \setlength{\tabcolsep}{0.2em} 
    {\scriptsize
    \renewcommand{\arraystretch}{1.2}
        \begin{tabular}{l||c|c|c|c|c}
                \hline
        $\z$ dim. & Enc. Recon. $\downarrow$ & KL & Rand. Recon. & BLEU $\uparrow$ & $|\params|$ \\ \hline \hline
        sVAE (32) + kl & 320.06 (0.04)   & 14.33 & 326.21 (0.01)  & 0.181 & 67M  \\
        sVAE (512) + kl & 329.2 (0.06)  & 7.09 & 331.35 (0.01)  & 0.139 & 70M  \\
        sVAE (1024) + kl & 334.41 (0.09)  & 5.52 & 335.83 (0.01) & 0.131 & 73M \\
        \hline
        sAE (32) &  293.75  &  & 487.45 (0.0) & 0.372 & 67M  \\
        sAE (512) & 222.34 &  & 375.38 (0.0)  & 0.624 & 70M  \\
        sAE (1024) & 326.66  &   & 374.31 (0.0) & 0.372  & 73M \\
        \hline
        sMIM (32) & 290.37 (0.01) & & 555.82 (0.0) & 0.387 & 67M \\
        sMIM (512) & 208.27 (0.0) & & 482.79 (0.01) & 0.664 & 70M \\
        sMIM (1024) & \textbf{205.81} (0.01) & & 475.16 (0.01) & \textbf{0.669} & 73M \\
        \hdashline[1pt/1pt]
        sMIM (1024)\textsuperscript{\textdagger} & \textbf{178.82} (0.0) &  &  & \textbf{0.682} & 179M \\
        \hline
        \end{tabular}
        }
    }
      \vspace*{-0.15cm}
    \caption{
    Reconstruction results for {\bf Yahoo Answers}, averaged over 10 runs.
    Models\textsuperscript{\textdagger} use extra training data
    (See Table \ref{tab:language-modelling-quantitative-yelp} for details).
    }
    \label{tab:language-modelling-quantitative-yahoo}
    \vspace*{-0.1cm}
\end{table}

\begin{table}[t]
    \centering
    \setlength{\tabcolsep}{0.2em} 
    {\scriptsize
    \renewcommand{\arraystretch}{1.2}
        \begin{tabular}{l||c|c|c|c}
                \hline
        $\z$ dim. & Enc. Recon. $\downarrow$ & KL & BLEU $\uparrow$ & $|\params|$ \\ \hline \hline
        sVAE (1024) + kl & 481.66 (0.1)  & 12.65   & 0.165 & 153 M  \\
        \hline
        sMIM (1024) & 329.89 (0.02) &   & 0.571 & 153 M  \\
        \hdashline[1pt/1pt]
        sMIM (1024)\textsuperscript{\textdagger} & \textbf{313.66} (0.01) & & \textbf{0.603} & 179M \\
        \hline
        \end{tabular}
    }
     \vspace*{-0.15cm}
    \caption{
    Reconstruction results for {\bf WT103} are averaged over 10 runs.
    Models\textsuperscript{\textdagger} use extra training data.
    The superior reconstruction results of sMIM hold for longer sentences.
    }
    \label{tab:language-modelling-quantitative-wiki103}
    \vspace*{-0.1cm}
\end{table}


Directly estimating the mutual information (MI) between the high-dimensional categorical observations $\x$ 
and the corresponding latent codes $\z$ is computationally expensive~\citep{Hjelm2018}.
Instead, we focus here on reconstruction, which is related to MI~\citep{poole2019variational}. 
We choose the reconstruction entropy (\textit{Enc.\ Recon.}), which is the negative expected log-probability of the decoder, given a sample from the corresponding posterior.
In addition, we show the reconstruction entropy when the latent code is sampled from a Gaussian prior (\textit{Rand.\ Recon.}). 
The Gaussian has 0 mean and standard deviation fitted to the latent codes (see Table\ \ref{tab:language-modelling-quantitative-entropy-fitted} in supplementary material).
When the latent code conveys little information to the decoder, we expect Enc.\ Recon.\ and Rand.\ Recon.\ to be similar.
When the decoder utilizes highly informative latent codes, we expect Rand.\ Recon.\ to be significantly larger (\ie, worse).
Finally we also show the 1-BLEU score, the fraction of words recovered in the sampled reconstruction.

Tables\ (\ref{tab:language-modelling-quantitative-ptb}-\ref{tab:language-modelling-quantitative-yahoo})
show reconstruction results for PTB, Yelp15, Yahoo Answers.  
For all datasets but PTB, VAE learning with KL annealing was more effective than standard VAE learning;
due to the small size of PTB, annealing overfit.
Model sMIM (1024)\textsuperscript{\textdagger} is trained on all datasets (\ie, PTB, Yahoo Answers, Yelp15 and WT103).
The BLEU score is computed between test sentences and their reconstructed samples (higher is better), 
and $|\params|$ indicates the number of parameters in each model.

Tables\ (\ref{tab:language-modelling-quantitative-ptb}-\ref{tab:language-modelling-quantitative-yahoo}) 
show that sMIM outperforms sVAE in reconstruction and BLEU score, and is comparable to sAE.
In addition, the reconstruction of sMIM and sAE improves with more latent dimensions, showing that more information is captured by the latent codes, 
whereas sVAE shows the opposite trend due to posterior collapse (\ie, encoder and random reconstructions are similar).
Notice that sAE is more susceptible to over-fitting, as is evident in Table\ \ref{tab:language-modelling-quantitative-yahoo}.
WT103 results in Table\ \ref{tab:language-modelling-quantitative-wiki103} show that the superior reconstruction of sMIM also holds with longer sentences.

In summary, sMIM shows improved performance with more expressive architecture (higher latent dimension here), 
similar to sAE, while sVAE deteriorates due to posterior collapse.
This suggests that sMIM could benefit from more powerful architectures like Transformers~\citep{Vaswani2017}, without the need for posterior collapse-mitigating heuristics.

\subsection{Posterior Collapse in VAE} \label{sec:nlp-posterior-collapse-ptb}

\begin{figure}[t]
    \centering
    \begin{subfigure}[t]{0.5\columnwidth}
        \centering
        \includegraphics[width=1.0\textwidth]{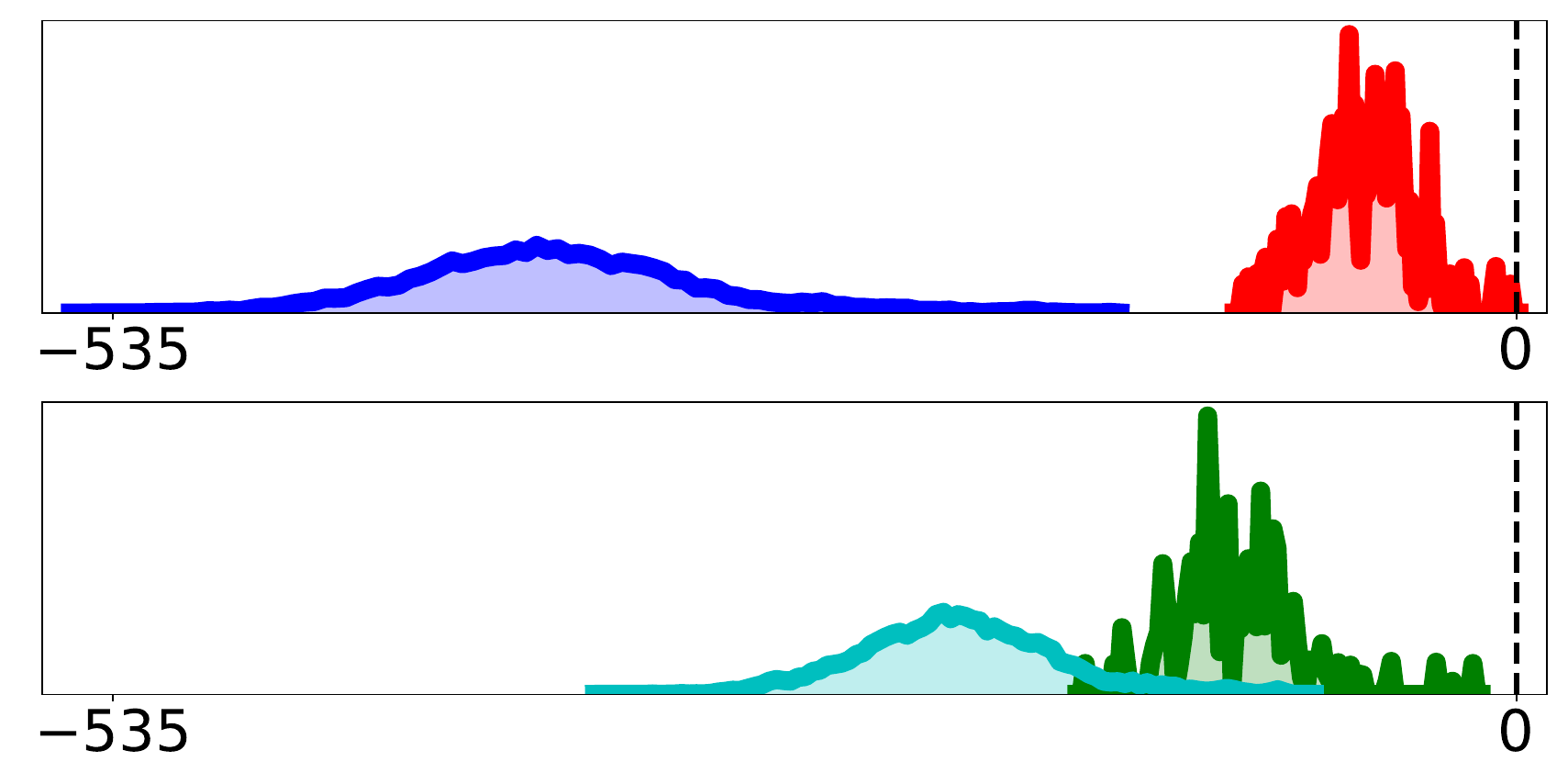}
        \caption{$\log \Mdec(\x_i|\z_j)$ histograms.}     
        \label{fig:nlp-dist-ptb-log_p_x_given_z}
    \end{subfigure}%
    \begin{subfigure}[t]{0.5\columnwidth}
        \centering
        \includegraphics[width=1.0\textwidth]{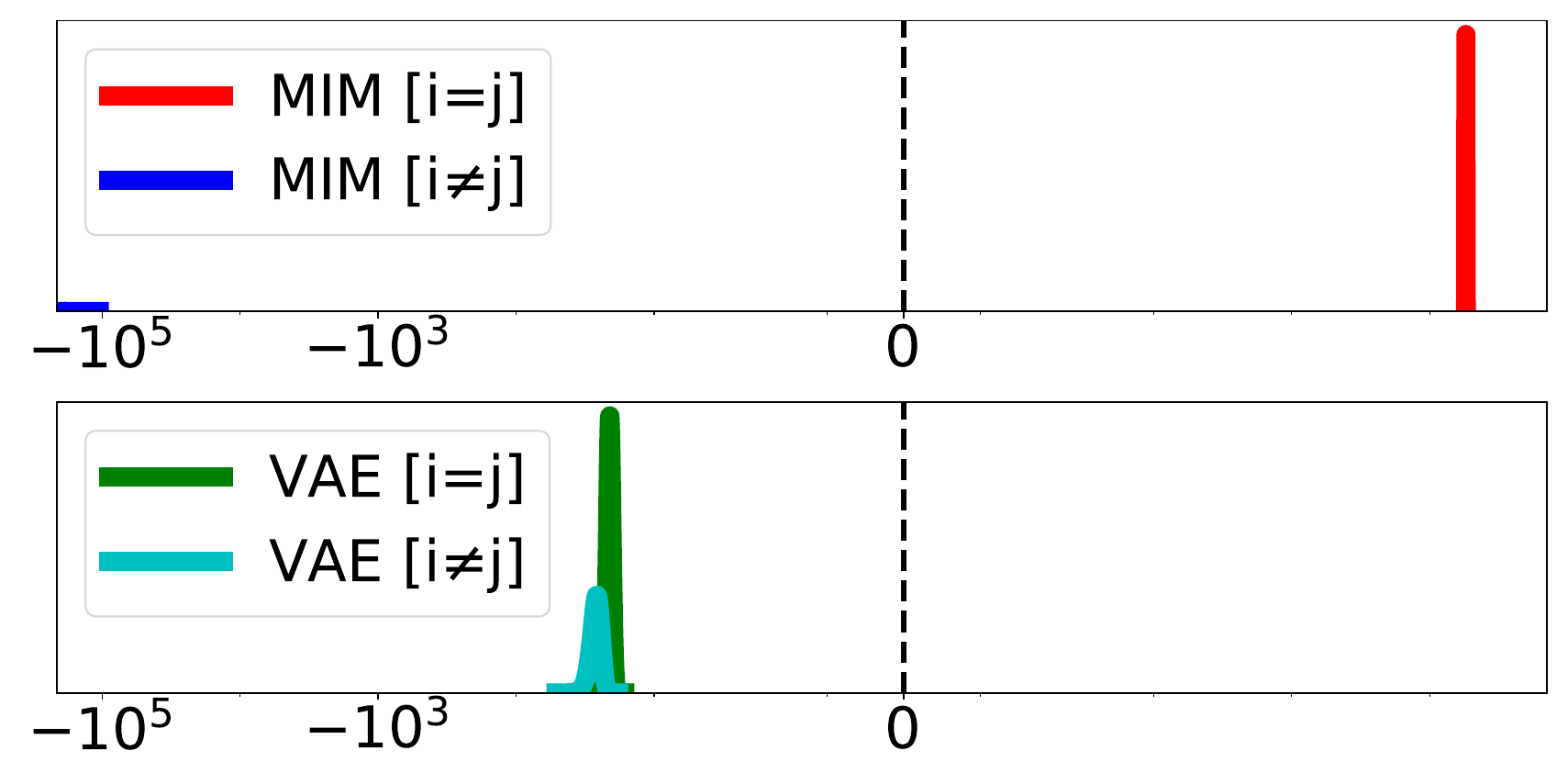}
        \caption{$\log \Menc(\z_i|\x_j)$ histograms.}     
        \label{fig:nlp-dist-ptb-log_q_z_given_x}
    \end{subfigure}
     \vspace*{-0.25cm}
    \caption{
    Histograms of log probabilities of test data for sMIM  and 
    sVAE trained on \textbf{PTB}: Overlap between  curves indicates
    potential for poor reconstruction of input sentences.
    (a) Histograms of $\log \Mdec(\x_i|\z_j)$ for $\z_j \sim \Menc(\z|\x_j)$ when $i=j$ 
    (same input), and when $i\neq j$ (when $\x_i$ is evaluated with the 
    decoder distribution from a latent code associated with a different input sentence).
    (b) Histograms of $\log \Menc(\z_i |\x_j)$ for  $\z_i\sim\Menc(\z |\x_i)$, 
    when conditioned on the same input $i = j$, or a different input $i\neq j$.
    \label{fig:nlp-dist-ptb}
    }
    \vspace*{-0.5cm}
\end{figure}

The performance gap between sMIM and sVAE  with identical architectures is due in part to posterior collapse in VAEs 
(\ie, optimization is likely to have a role too), 
where the encoder has high posterior variance over latent codes, and hence low mutual information
(cf.\ \citep{Zhao_Song_Ermon_2018,DBLP:journals/corr/abs-1711-00464}); it coincides with the KL divergence term in the 
usual ELBO approaching zero in some or all dimensions. 
In such cases, different sentences are mapped to similar regions of the latent space (see Fig.\ \ref{fig:nlp-dist-ptb}, bottom).

In contrast, given the high mutual information and reconstruction quality of sMIM, 
we only expect high decoding probability of a sentence when a latent code is sampled from
the corresponding posterior.
In other words, for sMIM, the posterior variances for different input sequences
are relatively small compared
to the distance between the posterior means (Fig.\ \ref{fig:nlp-dist-ptb}, top),
allowing for accurate reconstruction.


The issue of posterior collapse becomes harder to mitigate as the dimension of the latent space increases, because the decoder becomes more expressive.
As a consequence, language VAEs are typically limited to 32 dimensions or fewer (\eg, \citet{he2018lagging}), with only a few exceptions, such as \citet{Guu2017GeneratingSB} which opted for 128 dimensions in a very particular problem settings.  
On the other hand, sMIM can easily scale up the latent dimension without issue.

\subsection{Structure in the Latent Space} \label{sec:nlp-mim-vs-ae}

\begin{figure}[t]
    \vspace*{0.1cm}
    \centering
        \includegraphics[width=1.0\columnwidth]{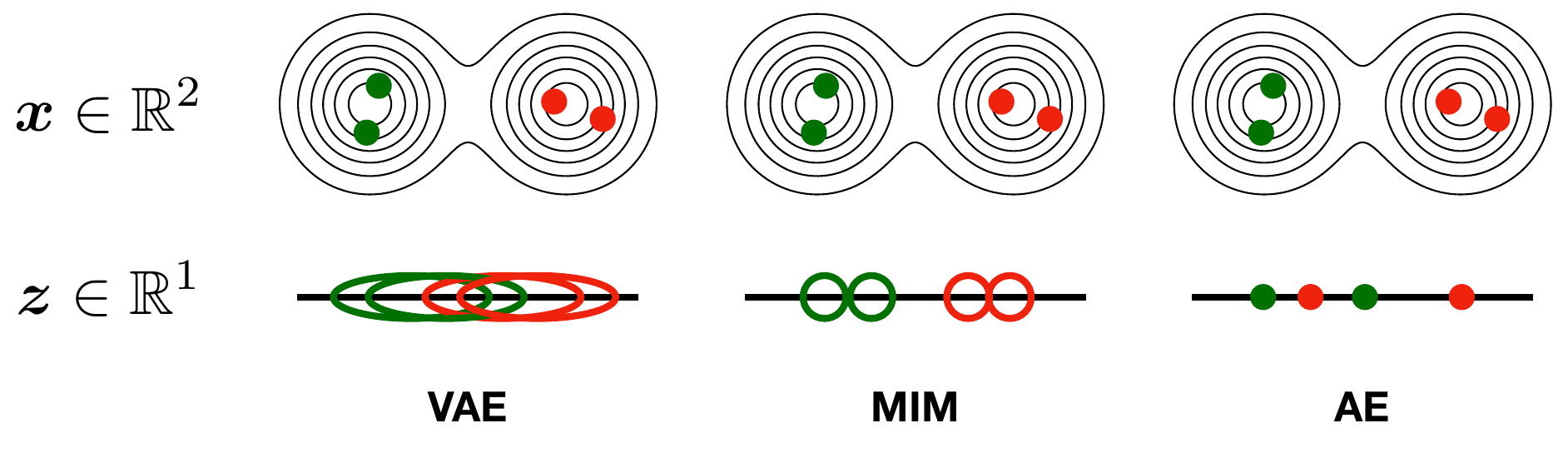}
        \vspace*{-0.8cm}
    \caption{
    \textbf{Top:} level-sets of a 2D data distribution (\ie, GMM with 2 modes).
    Red/green dots are samples.
    \textbf{Bottom:} the corresponding variance of the posterior, per sample, 
    for MIM and VAE, and the corresponding mapping for AE. 
    A semantically structured latent space will map latent samples from the same mode closer. A collapsed posterior (\ie, VAE), might mix nearby modes, 
    whereas MIM will have smaller posterior variance due to higher MI.
    A high entropy in the AE latent space might lead to non-semantic structure, where a perturbed red latent code might be reconstructed to the green mode, in contrast to MIM.
    \label{fig:latent-structure}
    }
\end{figure}

\begin{table}[t]
    \centering
    \scalebox{1.05}{
    \setlength{\tabcolsep}{0.2em} 
    {\scriptsize
    \renewcommand{\arraystretch}{1.2}
        \begin{tabular}{l||r|r|r|r}
        Dataset ($\z$ dim.) & \multicolumn{1}{c|}{sMIM} & \multicolumn{1}{c|}{$\mathcal{N}$} & \multicolumn{1}{c|}{sVAE} & \multicolumn{1}{c}{sAE}   \\ \hline \hline
        PTB (16D) & 11.54 [ 0.67 ]  & 22.7 & 23.04  [ 1.02 ]  & 35.95  [ 1.18 ]  \\
        PTB (128D) &  53.73 [ 0.55 ]  & 181.62 & 225.41 [ 1.24 ]  & 259.34 [ 1.26  ]  \\
        \hline
        Yelp15 (32D) & 32.22 [ 0.85 ]  & 45.4 & 50.01 [ 1.08 ]  & 73.03 [ 1.23 ]   \\
        Yelp15 (512D) &  186.18 [ 0.56 ] & 726.49 & 994.3 [ 1.37 ] & 917.0 [ 1.3 ] \\
        \hline
        Yahoo (32D) & 23.61 [ 0.73 ] & 45.4 & 44.45 [ 1.03 ] & 76.21 [ 1.26 ]   \\
        Yahoo (512D) &  155.47 [ 0.48 ] & 726.49 & 991.85 [ 1.37 ] & 1003.26 [ 1.35 ] \\
        \end{tabular}
        }
    }
    \vspace*{-0.1cm}
    \caption{
    Empirical entropy of the latent codes, estimated with a NN entropy estimator.
    For comparison, column $\mathcal{N}$ shows the entropy of a standard Normal in $\mathbb{R}^{d}$ of corresponding latent dimension.
    In brackets is the ratio of the NN entropy to the entropy of an isotropic Gaussian fit to the latent codes.
    Ratios below 1 indicate that the latent codes are more clustered than a Gaussian, suggesting the low entropy for sMIM is not a simple consequence of down-scaling the latent codes.
    }
    \label{tab:language-modelling-quantitative-entropy}
    \vspace*{-0.3cm}
\end{table}


Here, we explore the structure in the learned representation.
Table\ \ref{tab:language-modelling-quantitative-entropy} shows the empirical entropy, estimated using NN entropy estimator \citet{PhysRevE.69.066138}, of the latent codes for 
sMIM, sVAE, and sAE. Notice how the representation learned by sMIM has a significantly lower entropy.
We note that sVAE is regularized to match a Gaussian, which is known to introduce smoother structure into the learned representation (see discussion by \citet{bosc-vincent-2020-sequence}).

With sMIM, we propose the use of information entropy minimization as an alternative regularization, 
which introduces meaningful structure without suffering from posterior collapse (see schematic plot in Fig.\ \ref{fig:latent-structure}).
Interestingly, neuroscientists have proposed that entropy minimization is an organizing principle in neural representations and information processing in the brain. 
Entropy minimization is viewed as allowing the agent to learn better to predict likely events \citep{Friston2010, Barlow1972}, 
compression/redundancy elimination \citep{Barlow1961},
and in terms of efficiency in energy consumption \citep{10.3389/fncom.2019.00086}.

By scaling the latent codes to reduce the variance of the aggregate posterior one can trivially reduce entropy with no benefit in terms of latent structure.
To test whether this might be the case, we also fit an isotropic Gaussian to the latent codes
(see Table\ \ref{tab:language-modelling-quantitative-entropy-fitted} in supplementary materials), 
and show the ratio between the NN entropy and the fitted entropy in brackets.
A ratio smaller than 1 suggests that the empirical entropy is more clustered than a Gaussian.
Table\ \ref{tab:language-modelling-quantitative-entropy} clearly shows that the lower empirical 
entropy cannot be explained by the scaling alone.
We attribute the gap between the fitted and empirical entropies to clustering in the latent space,  
which MIM learning empirically demonstrates \citep{2019arXiv191003175L}.

\subsection{Reconstruction, Interpolation, and Perturbation} \label{sec:nlp-recon}

\begin{table}[t]
    \centering
    \setlength{\tabcolsep}{0.5em} 
    {\scriptsize
    \renewcommand{\arraystretch}{1.2}
        \begin{tabular}{p{8cm}}
        \textbf{5 stars $\rightarrow$ 1 star} \\ \hline \hline
        \rowcolor{Gray}
\textsc{<BOT>} awesome food , just awesome ! top notch beer selection . great staff . beer garden is great setting . \\
\hline
\textbullet ~ awesome food , just top notch ! great beer selection . staff has great craft beer . top notch is that . \textsc{<EOT>} \\
\textbullet ~ awesome food ! just kidding , beer selection is great . staff has trained knowledge on top .  \textsc{<EOT>} \\
\textbullet ~ cleanliness is awesome ! not only on their game , food . server was polite his hand sanitizer outside .  \textsc{<EOT>} \\
\textbullet ~ cleanliness is not on their patio . server was outside , kept running his hand sanitizer his hand .  \textsc{<EOT>} \\
\hline
        \rowcolor{Gray}
 \textsc{<BOT>} cleanliness is not on their radar . outside patio was filthy , server kept running his hand thru his hair . \\ \hline \hline
\end{tabular}
    }
        \vspace*{-0.2cm}
    \caption{
    Interpolation results between latent codes of input sentences (with gray) from \textbf{Yelp15} for sMIM (1024).
    }
    \label{tab:language-modelling-ptb-qualitative-interpolation-mim-2}
    \vspace*{-0.15cm}
\end{table}

\begin{table}[t]
    \centering
    \setlength{\tabcolsep}{0.5em} 
    {\scriptsize
    \renewcommand{\arraystretch}{1.2}
        \begin{tabular}{l p{7cm} }
        \hline \hline
        \rowcolor{Gray}
(D) & \textsc{<BOT>} the company did n't break out its fourth-quarter results  \\
        \hline 
(M) & the company did n't break out its results \textsc{<EOT>} \\
(R) & the company did n't break out its fourth-quarter results \textsc{<EOT>} \\
(P) & the company did n't accurately out its results \textsc{<EOT>} \\

    \end{tabular}
    }
    \vspace*{-0.2cm}
    \caption{
    Reconstruction results for sMIM (512) model trained on \textbf{PTB}.
    We denote:
    (D) Data sample; 
    (M) Mean (latent) reconstruction;
    (R) Reconstruction;
    (P) Perturbed (latent) reconstruction.
    }
    \label{tab:language-modelling-quantitative-ptb-recon}
    \vspace*{-0.2cm}
\end{table}

We further probe the structure in the learned representation, demonstrating that
sMIM learns a dense, meaningful latent space. We present latent interpolation results in
Table\ \ref{tab:language-modelling-ptb-qualitative-interpolation-mim-2}
for samples (\ie, reviews) with the different ratings from Yelp5. 
Interpolation entails sampling $\x \sim \Mdec(\x| \z_\alpha)$ where $\z_\alpha$ is interpolated at equispaced points  
between two randomly sampled latent codes, $\z_i \sim \Menc(\z|\x_i)$, and $\z_j \sim \Menc(\z|\x_j)$.

Next we show reconstruction, and perturbation
results for for sMIM (512) trained on PTB. 
Table\ \ref{tab:language-modelling-quantitative-ptb-recon} shows four sentences:
(D) the input sentence;
(M) the mean reconstruction given the posterior mean $z$;
(R) a reconstruction given a random sample $z$ from the posterior;
and 
(P) a {\em perturbed reconstruction}, given a sample $z$ from a Gaussian
distribution with 10 times the posterior standard deviation.
The high mutual information learned by sMIM leads to good reconstruction, as clear in (M) and (R).
sMIM also exhibits good clustering in the latent space, shown here by the similarity of (R) and (P).

\subsection{Question-Answering} 
\label{sec:nlp-question-answering-yahoo}

\begin{table}[t]
    \centering
    \setlength{\tabcolsep}{0.2em} 
    {\footnotesize
    \renewcommand{\arraystretch}{1.2}
\begin{tabular}{l||ccc}
        \hline
    Model & P@1 $\uparrow$ & MRR  $\uparrow$  \\
    \hline \hline
    AP-CNN \citep{DBLP:journals/corr/SantosTXZ16} & 0.560 & 0.726   \\
    AP-BiLSTM \citep{DBLP:journals/corr/SantosTXZ16} & 0.568 & 0.731  \\
    HyperQA \cite{DBLP:journals/corr/TayLH17a} &  \textbf{0.683} & 0.801   \\
    \hline
    sAE (32)  & 0.386 & 0.656 \\
    sAE (512)  & 0.579 & 0.814 \\
    sAE (1024)  & 0.519 & 0.767 \\
    \hline
    sVAE (32) + kl \textsuperscript{*} & 0.531 & 0.776 \\
    sVAE (512) + kl \textsuperscript{*} & 0.494 & 0.747 \\
    sVAE (1024) + kl \textsuperscript{*} & 0.548 & 0.79 \\
    \hline
    sMIM (32) \textsuperscript{\textdaggerdbl} & 0.558  & 0.737 \\
    sMIM (512) \textsuperscript{\textdaggerdbl} & \textbf{0.683} & \textbf{0.818} \\
    sMIM (1024) \textsuperscript{\textdaggerdbl} & 0.651 & 0.8 \\
    \hdashline[1pt/1pt]
    sAE (1024)  \textsuperscript{\textdagger} & 0.574 & 0.812 \\
    sVAE (1024) \textsuperscript{\textdagger}\textsuperscript{*} & 0.339  & 0.616 \\
    sMIM (1024) \textsuperscript{\textdagger} \textsuperscript{\textdaggerdbl} & \textbf{0.757} & \textbf{0.863} \\
        \hline
\end{tabular}
    }
        \vspace*{-0.1cm}
    \caption{
    {\bf YahooCQA} results for sMIM, AE, and single-task models.
    Results\textsuperscript{\textdaggerdbl} are averaged over 10 runs (stdev $< 0.002$).
    sMIM (1024)\textsuperscript{\textdagger} is pre-trained on Everything dataset. 
    sVAE\textsuperscript{*} results are based on the mean of the posterior, rather the sample \citep{bosc-vincent-2020-sequence}. 
    $P@1$ and MRR are defined in Sec.\ \ref{sec:nlp-question-answering-yahoo}.
    }
    \label{tab:question-answering-yahoocqa}
    \vspace*{-0.1cm}
\end{table}


\begin{table}[t]
    \centering
    \setlength{\tabcolsep}{0.5em} 
    {\scriptsize
    \renewcommand{\arraystretch}{1.2}
        \begin{tabular}{p{8cm}}
         \hline
        \rowcolor{Gray}
Q: \textsc{<BOT>} my brother is geting out on parole from navy jail where can i find a parole office in our area \textsc{<UNK>} , \textsc{<UNK>} ? \\
\hline
A: you can find out the county jail , or call your local police station . \textsc{<EOT>} \\
\hline
         \hline
        \rowcolor{Gray}
Q: \textsc{<BOT>} what continent has most deserts ? \\
\hline
A: the most notable is in the netherlands . \textsc{<EOT>} \\
\hline
         \hline
        \rowcolor{Gray}
Q: \textsc{<BOT>} how do u clear the history in the search field ? \\
\hline
A:  u can find it in the search bar . \textsc{<EOT>} \\
\hline
         \hline
        \rowcolor{Gray}
Q: \textsc{<BOT>} what is the best question to ask ? \\
\hline
A: ask yourself ! \textsc{<EOT>} \\
\hline
         \hline
        \rowcolor{Gray}
Q: \textsc{<BOT>} need to find somewhere to sale baseball cards . ? \\
\hline
A: ebay \textsc{<EOT>} \\
\hline
         \hline
\end{tabular}
    }
        \vspace*{-0.1cm}
    \caption{
    Sampled answers from \textbf{Yahoo Answers} sMIM (1024).
    }
    \label{tab:question-answering-samples}
    \vspace*{-0.1cm}
\end{table}

So far we have discussed abstract aspects of representations learned by sMIM, such as high mutual information, and low marginal entropy.
To demonstrate the benefits of representations learned by sMIM, we consider a downstream task in which sMIM is 
pre-trained on Yahoo Answers, then used for question-answering on YahooCQA 
\citep{DBLP:conf/sigir/TayPLH17}, with no fine-tuning. 
YahooCQA comprises questions and 3-5 answers, where the first answer is from Yahoo Answer, 
and the additional answers are wrong.
Let $Q_i$ denote the $i^{th}$ question, and let $ \{A_i^k\}_{k=1}^{K_i}$ be the $K_i$ 
corresponding answers, ordered such that $A_i^k$ has rank $k$.
To match the format of QA pairs in Yahoo Answers, we compose question-answer 
pair  $Q_i^k$ by concatenating $Q_i$, "?", and $A_i^k$.

For question-answering with sMIM we use the following procedure:
For each question-answer we sample $\z_i^k \sim \Menc(\z|Q_i^k)$, and a corresponding 
$\z_i^{unk} \sim \Menc(\z|Q_i^{unk})$ where $Q_i^{unk}$ is simply $Q_i$ concatenated 
with "?" and a sequence of <unk> tokens to represent the $|A_i^k|$ unknown words of the answer.
We then rank question-answer pairs according to the score
$S_i^k = ||\z_i^{unk} - \z_i^k|| / \sigma_i^{k,unk}$
where $\sigma_i^{k,unk}$ is the standard deviation of $\Menc(\z|Q_i^{unk})$. 
In other words, we rank each question-answer pair according to the normalized distance between  
the latent code of the question with, and without, the answer. This score is similar 
to  $\log \Menc(\z_i^k|Q_i^{unk})$, but without taking the log standard deviation into account. 

As is common in the literature, Table\ \ref{tab:question-answering-yahoocqa} quantifies test performance 
using average precision 
($P@1 \!=\! \frac{1}{N} \sum_i\! \mathbbm{1}{(rank(A_i^1) = 1)}$), 
and Mean Reciprocal Ranking ($MRR = \frac{1}{N} \sum_i \frac{1}{rank(A_i^1)}$).
As baselines, we consider best performing single-task models trained directly on YahooCQA \citep{DBLP:journals/corr/SantosTXZ16,DBLP:journals/corr/TayLH17a}.
Interestingly, sMIM (512), pre-trained on Yahoo Answers, exhibits state-of-the-art 
performance compared to these baselines.
For an even larger sMIM model, pre-trained on all of PTB, Yahoo Answers, Yelp15
and WT103, the question-answering performance of sMIM is even better
(last row of Table\ \ref{tab:question-answering-yahoocqa}).

The results for sVAE are based on the mean of the posterior rather than a random sample.
This is a common heuristic in the NLP literature which has been proven useful for downstream tasks, 
but is problematic when considering the generative process which relies on the sample rather than the mean \citep{bosc-vincent-2020-sequence}. 
Finally, as another point of comparison, we repeated the experiment with a 
deterministic sAE model (with $\sigma_i^{k,unk} = 1$).
In this case performance drops, especially average precision, indicating that 
the latent representations are not as meaningfully structured.

Importantly, sMIM can generate novel answers rather than simply ranking a given set of alternatives.
To this end, we sample $\z_i^{unk} \sim \Menc(\z_i^k|Q_i^{unk})$, as described above,
followed by modified reconstruction $\widehat{Q}_i \sim \Mdec(\x|\z_i^{unk})$. 
We modify the sampling procedure to be greedy (\ie, top 1 token), and prevent 
the model from sampling the "\textsc{<UNK>}" token.
We consider all words past the first "?" as the answer. (We also removed HTML tags (\eg, "<br>").)
Table\ \ref{tab:question-answering-samples} gives several selected answers.
The examples were chosen to be short, and with appropriate (non-offensive) content.
We note that we did not use any common techniques to manipulate the decoding distribution,
such as beam search, Nucleus sampling \citep{DBLP:journals/corr/abs-1904-09751}, or sampling with temperature \citep{ACKLEY1985147}.
To the best of our knowledge, sMIM is the current state-o-the-art for a {\em single-task} model for YahooCQA, 
despite having simpler architecture and training procedure when compared to competing models.


\section{Conclusions}

This paper introduces sMIM, a new probabilistic auto-encoder for language modeling,
trained with A-MIM learning.  
In particular, sMIM avoids posterior collapse, a challenging problem with VAEs applied to language,
which enables the use of larger latent dimensions compared to sVAE, by orders of magnitude.
While the reconstruction error is comparable to sAE, sMIM learns a latent representation with semantic structure, similar to sVAE, 
allowing for interpolation, perturbation, and sampling. In this sense, it achieves the best of both worlds: a semantically meaningful representation with high information content.
We also use the structured latent representation for a downstream question-answering task on YahooCQA with state-of-the-art results.
Importantly, the proposed framework has no hyper-parameters in the loss, greatly simplifying the optimization procedure.
In addition, sMIM benefits from a more expressive architecture, in contrast to sVAE, and demonstrates reduced susceptibility to over-fitting, compared to sAE. 
In future work, we will apply sMIM to more contemporary and powerful architectures like the Transformer.



\bibliography{paper}
\bibliographystyle{icml2021}

\newpage
\onecolumn
\appendix
\section{Distribution of Sentence Lengths} 

\begin{figure}[h]
    \vspace*{-0.25cm}
    \centering
    \setlength{\tabcolsep}{0pt}
    \begin{tabular}{*4{>{\centering\arraybackslash}m{0.25\textwidth}}}
     \includegraphics[width=0.23\columnwidth]{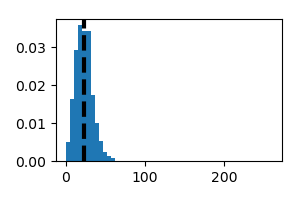}
     &
     \includegraphics[width=0.23\columnwidth]{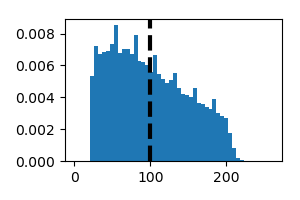}
     &
     \includegraphics[width=0.23\columnwidth]{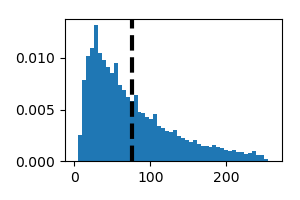}
     &
     \includegraphics[width=0.23\columnwidth]{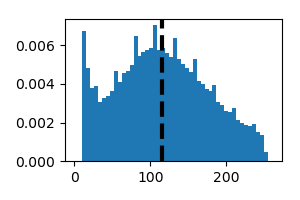}
    \\
    (a) PTB  & (b) Yelp15 & (c) Yahoo Answers & WT103 
    \end{tabular}
    \vspace*{-0.25cm}
    \caption{
    Here we present histograms of sentence lengths per dataset. 
    The dashed line is the average sentence length.
    }
    \label{fig:nlp-datasets-hist}
\end{figure}

Fig.\ \ref{fig:nlp-datasets-hist} shows histograms of sentence lengths. 
Notice that PTB sentences are significantly shorter that other datasets.
As a result, sMIM is somewhat better able to learn a representation that is well suited for reconstruction.
Other datasets, with longer sentences, are more challenging, especially with the simple architecture used here (\ie, 1 later GRU).
We believe that implementing sMIM with an architecture that better handles long-term dependencies
(\eg, transformers) might help.

\FloatBarrier

\section{Comparison of Reconstruction in MIM and VAE}

\begin{figure}[th]
    \centering
    \setlength{\tabcolsep}{0pt}
    \begin{tabular}{m{0.1\textwidth} *3{>{\centering\arraybackslash}m{0.3\textwidth}}}
     sMIM &
     \includegraphics[width=0.23\columnwidth]{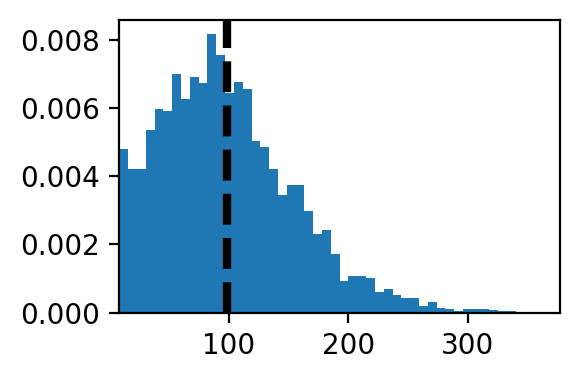}
     &
     \includegraphics[width=0.23\columnwidth]{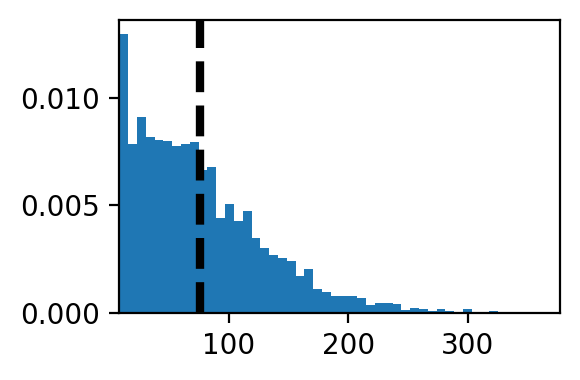}
     &
     \includegraphics[width=0.23\columnwidth]{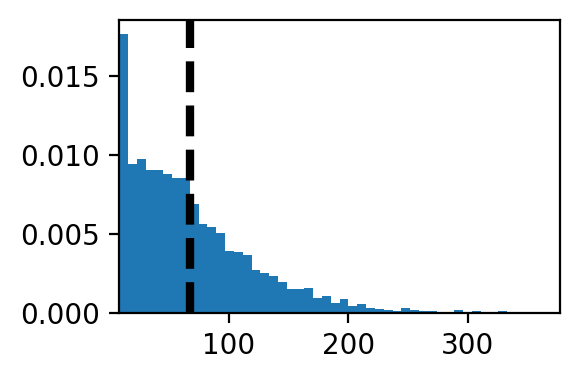}
    \\
     sVAE &
     \includegraphics[width=0.23\columnwidth]{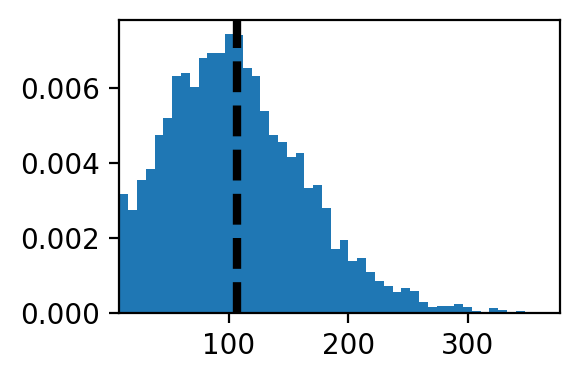}
     &
     \includegraphics[width=0.23\columnwidth]{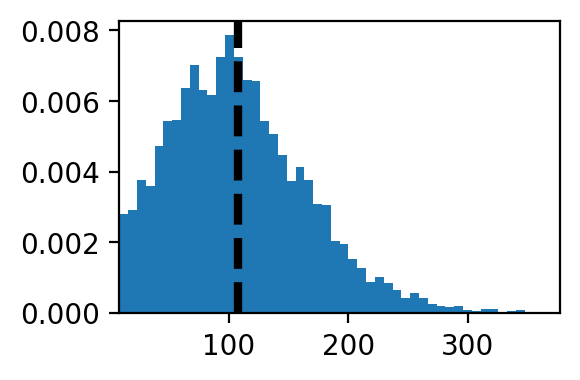}
     &
     \includegraphics[width=0.23\columnwidth]{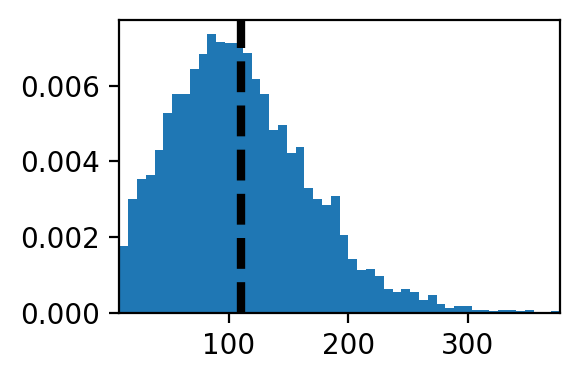}
     \\
     & $\z \in \mathbb{R}^{16}$ & $\z \in \mathbb{R}^{128}$ & $\z \in \mathbb{R}^{512}$ 
    \end{tabular}
        \vspace*{-0.25cm}
    \caption{
     Histograms of reconstruction for sMIM and sVAE versus latent dimension for \textbf{PTB}.
    Dashed black line is the mean.
    }
    \label{fig:nlp-nll-hist-ptb}
\end{figure}

\begin{figure}[th]
    \centering
    \setlength{\tabcolsep}{0pt}
    \begin{tabular}{m{0.1\textwidth} *3{>{\centering\arraybackslash}m{0.3\textwidth}}}
     sMIM &
     \includegraphics[width=0.23\columnwidth]{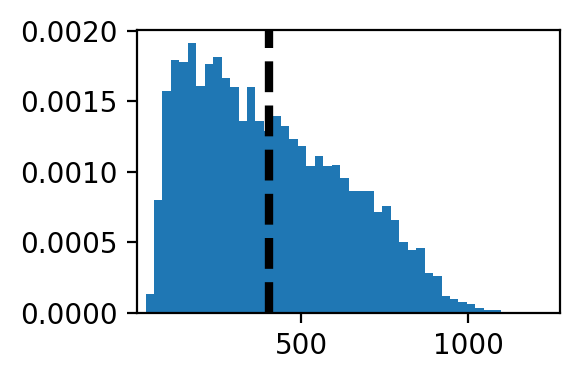}
     &
     \includegraphics[width=0.23\columnwidth]{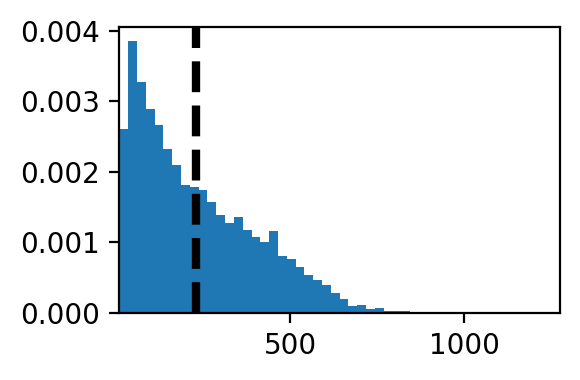}
     &
     \includegraphics[width=0.23\columnwidth]{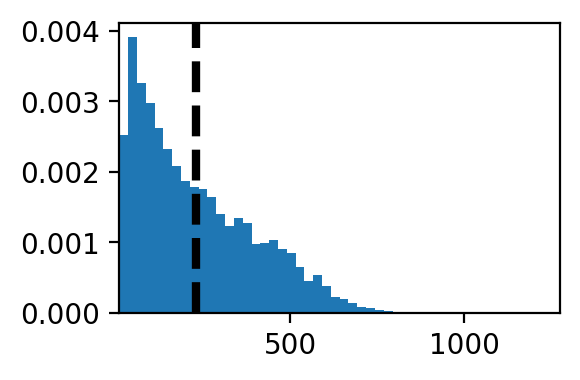}
    \\
     sVAE + kl &
     \includegraphics[width=0.23\columnwidth]{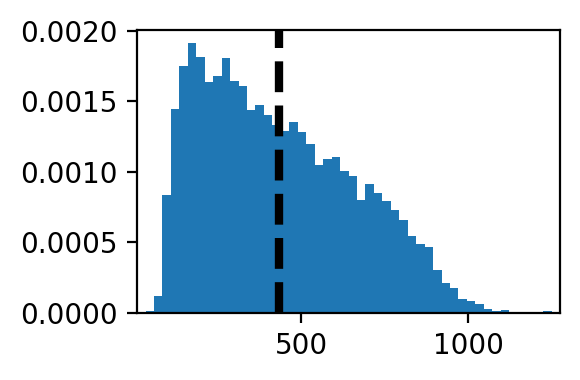}
     &
     \includegraphics[width=0.23\columnwidth]{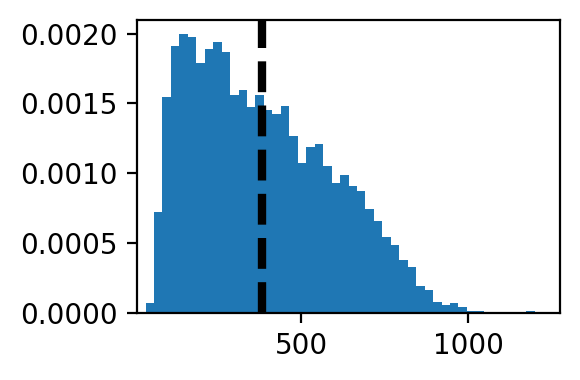}
     &
     \includegraphics[width=0.23\columnwidth]{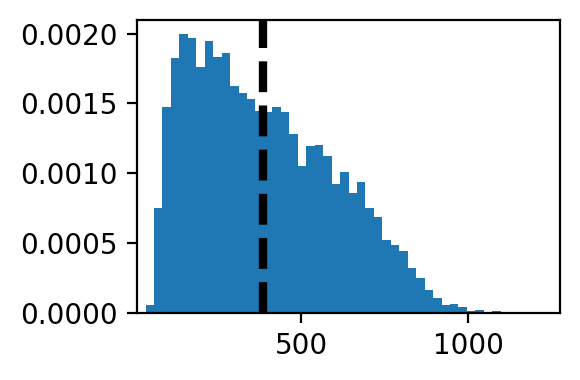}
     \\
     & $\z \in \mathbb{R}^{32}$ & $\z \in \mathbb{R}^{512}$ & $\z \in \mathbb{R}^{1024}$ 
    \end{tabular}
        \vspace*{-0.25cm}
    \caption{  
    Histograms of reconstruction for sMIM and sVAE versus latent dimension for \textbf{Yelp15}.
    Dashed black line is the mean.
    }
    \label{fig:nlp-nll-hist-yelp}
\end{figure}

\begin{figure}[th]
    \centering
    \setlength{\tabcolsep}{0pt}
    \begin{tabular}{m{0.1\textwidth} *3{>{\centering\arraybackslash}m{0.3\textwidth}}}
     sMIM &
     \includegraphics[width=0.23\columnwidth]{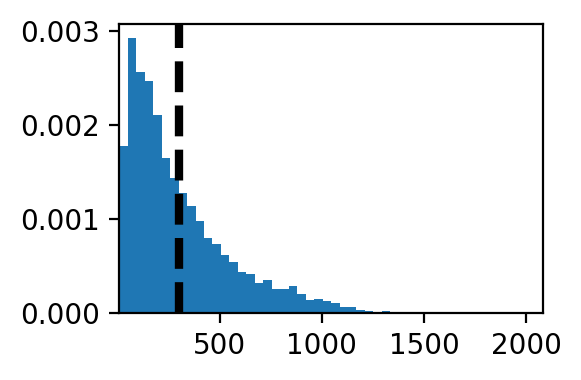}
     &
     \includegraphics[width=0.23\columnwidth]{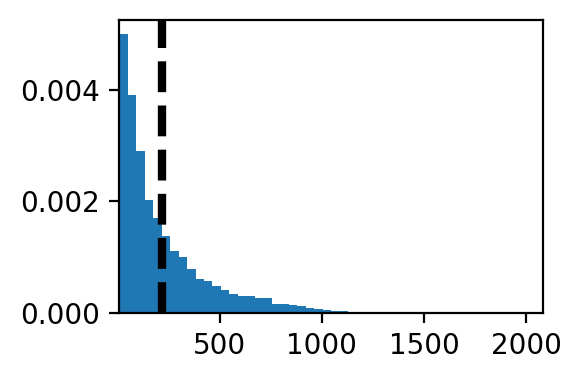}
     &
     \includegraphics[width=0.23\columnwidth]{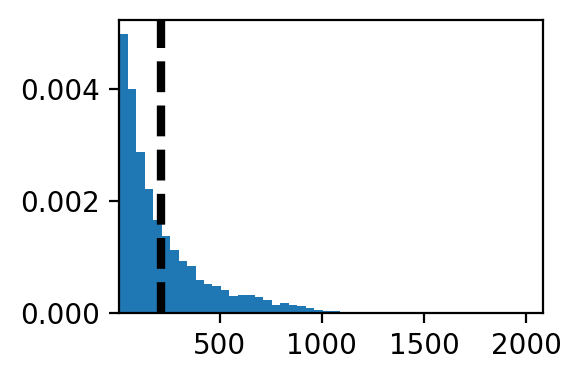}
    \\
     sVAE + kl &
     \includegraphics[width=0.23\columnwidth]{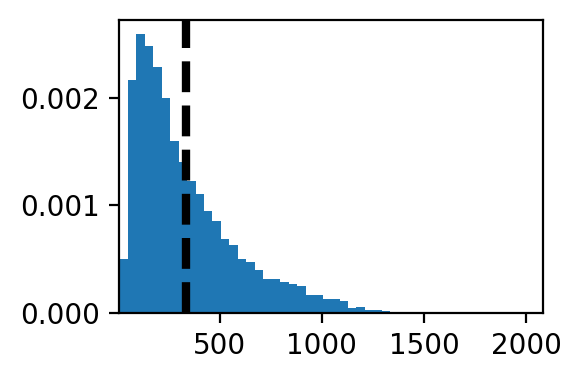}
     &
     \includegraphics[width=0.23\columnwidth]{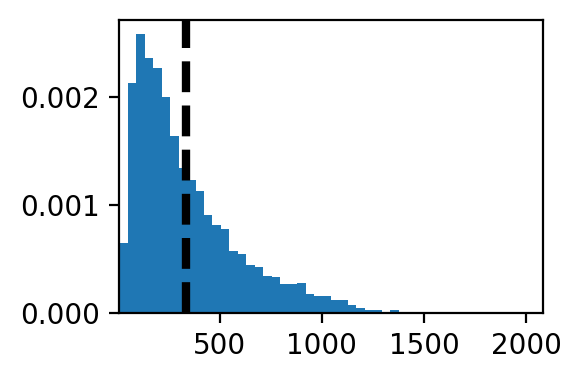}
     &
     \includegraphics[width=0.23\columnwidth]{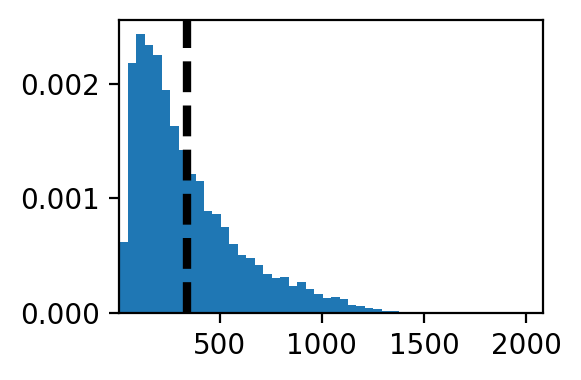}
     \\
     & $\z \in \mathbb{R}^{32}$ & $\z \in \mathbb{R}^{512}$ & $\z \in \mathbb{R}^{1024}$ 
    \end{tabular}
        \vspace*{-0.25cm}
    \caption{
  Histograms of reconstruction for sMIM and sVAE versus latent dimension for \textbf{Yahoo Answers}.
    }
    \label{fig:nlp-nll-hist-yahoo}
\end{figure}

Figures \ref{fig:nlp-nll-hist-ptb}-\ref{fig:nlp-nll-hist-yahoo} depict histograms of reconstruction values for sentences, for sVAE and sMIM with different latent dimensions.
While a less expressive sMIM behaves much like sVAE, the difference is clearer as the expressiveness of the model increases. 
Here, sVAE does not appear to effectively use the increased expressiveness for better modelling.
We hypothesize that the added sVAE expressiveness is used to better match the posterior 
to the prior, resulting in posterior collapse. sMIM uses the increased expressiveness to 
increase mutual information.

\FloatBarrier

\section{Empirical Latent Entropy} 

\begin{table}[t]
    \centering
    \setlength{\tabcolsep}{0.2em} 
    {\scriptsize
    \renewcommand{\arraystretch}{1.2}
        \begin{tabular}{l||cccc}
        Dataset ($\z$ dim.) & sMIM & $\mathcal{N}$ & sVAE & sAE   \\ \hline \hline
        PTB (16D) & [ 17.22 / 0.5 ] & 22.7 &  [ 22.51 / 0.97 ] &  [ 30.49 / 2.64 ]  \\
        PTB (128D) &  [ 97.39 / 0.26 ] & 181.62 & [ 181.29 / 0.99 ] & [ 206.22 / 1.46 ] \\
        \hline
        Yelp15 (32D) & [ 38.06 / 0.63 ] & 45.4 &  [46.31 / 1.05] & [ 59.16 / 0.99 ]  \\
        Yelp15 (512D) &  [ 333.45 / 0.21 ] & 726.49  & [ 726.15 / 0.99 ] & [ 705.14 / 0.91 ] \\
        \hline
        Yahoo (32D) &  [ 32.13 / 0.43 ] & 45.4 & [43.3 / 0.87 ] & [ 60.45 / 2.56 ]  \\
        Yahoo (512D) &  [ 326.17 / 0.2 ] & 726.49  &  [ 724.75 / 0.99 ] & [ 744.24 / 1.07 ] \\
        \end{tabular}
    }
    \vspace*{-0.1cm}
    \caption{
    In brackets is the entropy of an isotropic Gaussian fitted to the latent codes, and the corresponding average standard deviation [ entropy / stdev ].
    For comparison, column $\mathcal{N}$ shows the entropy of a standard Normal in $\mathbb{R}^{d}$ of a corresponding latent dimension.
    Our goal here is to rule out simple down-scaling as the cause for the low entropy in sMIM.
    }
    \label{tab:language-modelling-quantitative-entropy-fitted}
    \vspace*{-0.1cm}
\end{table}

Table\ \ref{tab:language-modelling-quantitative-entropy-fitted} provides the entropy of an isotropic Gaussian that is fitted to the latent codes, and the standard deviation of the fitted Gaussian [entropy / stdev]. The values are used to compute the ratio presented in the main paper.

\FloatBarrier

\section{Additional Results}

\subsection{Reconstruction}

\begin{table}[th]
    \centering
    \setlength{\tabcolsep}{0.5em} 
    {\scriptsize
    \renewcommand{\arraystretch}{1.2}
        \begin{tabular}{l| p{8cm}|p{8cm} }
        \hline \hline
        & sMIM (512) & sMIM (1024) \textsuperscript{\textdagger} \\
        \hline \hline
        \rowcolor{Gray}
(D) & \multicolumn{2}{l}{\textsc{<BOT>} there was no panic }  \\
        \hline 
\hdashline[1pt/1pt]
(M) & there was no panic \textsc{<EOT>} & there was no panic  \textsc{<EOT>} \\
\hdashline[1pt/1pt]
(R) & there was no orders  \textsc{<EOT>} & there was no panic \textsc{<EOT>} \\
\hdashline[1pt/1pt]
(P) & there was no panic \textsc{<EOT>} & there was no shortage panic \textsc{<EOT>} \\
\hdashline[1pt/1pt]
(AE) & there was no panic \textsc{<EOT>} &  \\
        \hline 
        \hline 
        \rowcolor{Gray}
(D) & \multicolumn{2}{l}{\textsc{<BOT>} the company did n't break out its fourth-quarter results}  \\
        \hline 
\hdashline[1pt/1pt]
(M) & the company did n't break out its fourth-quarter results  \textsc{<EOT>} & the company did n't break out its results results \textsc{<EOT>} \\
\hdashline[1pt/1pt]
(R) & the company did n't break out its results \textsc{<EOT>} & the company did n't break out its results \textsc{<EOT>} \\
\hdashline[1pt/1pt]
(P) & the company did n't break out its fourth-quarter results \textsc{<EOT>} & the company did n't break out its results results \textsc{<EOT>} \\
\hdashline[1pt/1pt]
(AE) & the company did n't break out results \textsc{<EOT>} &  \\
        \hline 
        \hline 
        \rowcolor{Gray}
(D) & \multicolumn{2}{l}{\textsc{<BOT>} it had planned a strike vote for next sunday but that has been pushed back indefinitely}  \\
        \hline 
\hdashline[1pt/1pt]
(M) & it had a weakening for promotional planned but that has pushed aside back but so far away \textsc{<EOT>} & it had planned planned a planned for next week but that continues has been pushed back pushed \textsc{<EOT>} \\
\hdashline[1pt/1pt]
(R) & it had a planned strike for energy gifts but so that has planned airlines but block after six months \textsc{<EOT>} & it had planned a strike planned for next sunday but that has been pushed back culmination pushed \textsc{<EOT>} \\
\hdashline[1pt/1pt]
(P) & it had a strike with stateswest airlines but so that it has slashed its spending but so far said he would be subject by far \textsc{<EOT>} & it had planned a strike for hardcore but has been pushed every year that leaves back \textsc{<EOT>} \\
\hdashline[1pt/1pt]
(AE) & it had been a five-year vote but for a week that drilling humana strike back back has planned back \textsc{<EOT>} &  \\
    \end{tabular}
    }
    \caption{
    Reconstruction results for models trained on \textbf{PTB}.
    We denote:
    (D) Data sample; 
    (M) Mean (latent) reconstruction;
    (R) Reconstruction;
    (P) Perturbed (latent) reconstruction;
    (AE) Reconstruction of AE.
    }
    \label{tab:nlp-recon-ptb-1}
\end{table}

\begin{table}[th]
    \centering
    \setlength{\tabcolsep}{0.5em} 
    {\scriptsize
    \renewcommand{\arraystretch}{1.2}
        \begin{tabular}{l| p{8cm}|p{8cm} }
        \hline \hline
        & sMIM (1024) & sMIM (1024) \textsuperscript{\textdagger} \\
        \hline \hline
        \rowcolor{Gray}
(D) & \multicolumn{2}{l}{\textbf{(3 stars)}  \textsc{<BOT>} decent price . fast . ok staff ... but it is fast food so i ca n't rate any higher than 3 .}  \\
        \hline 
\hdashline[1pt/1pt]
(M) & decent italians . fast . price ok ... but it is higher than any other fast food i ca n't rate so higher rate jusqu . \textsc{<EOT>} & decent oxtail . ok . fast price ... but staff it is so fast i ca n't rate any food 3 . \textsc{<EOT>} \\
\hdashline[1pt/1pt]
(R) & decent price . superior . decent staff ... but ok fast food is n't so it i ' d rate higher any higher quality than 3 . \textsc{<EOT>} & decent price . fast staff . fast ok ... but it is so fast food i rate 3 higher than any . \textsc{<EOT>} \\
\hdashline[1pt/1pt]
(P) & decent price . ok . fast food ... but it is ok . so i ca n't rate any higher rate as fast food is marginal . \textsc{<EOT>} & decent price . fast . wu ... fast food ! but it staff so ok i ca n't rate 3 stars . . \textsc{<EOT>} \\
\hdashline[1pt/1pt]
(AE) & decent price . fast staff . ok ... but it is fast food so i ca n't rate any rate than 3 . \textsc{<EOT>} &  \\
        \hline 
        \hline 
        \rowcolor{Gray}
(D) & \multicolumn{2}{l}{\textbf{(4 stars)} \textsc{<BOT>} excellent wings . great service . 100 \% smoked wings . great flavor . big meaty . i will definitely be back . okra is great too . }  \\
        \hline 
\hdashline[1pt/1pt]
(M) & excellent wings . great service . 100 \% wings . big meaty wings . great flavor . i definitely will be back . lake is great too . \textsc{<EOT>} & excellent service . great wings . 100 \% superior . great flavor . great fries . definitely will be back . i had too big fat . \textsc{<EOT>} \\
\hdashline[1pt/1pt]
(R) & excellent wings . great service . 100 \% wings . wings flavor . definitely great . 100 \% . i will be back . \textsc{<EOT>} & excellent service . great flavor . 100 \% wings . excellent . great big guts . definitely will be back from . i had great wings . \textsc{<EOT>} \\
\hdashline[1pt/1pt]
(P) & excellent wings . great service . wings flavours wings . 100 \% big . mmmmm overwhelmed . i ' m definitely hooked . bye disgusted is great but will be back . i definitely go . \textsc{<EOT>} & great burger . excellent service . 100 \% fat bowls . great carnitas . great flavor . i will definitely be back . i avoid too late . \textsc{<EOT>} \\
\hdashline[1pt/1pt]
(AE) & excellent excellent . great service . 100 \% wings . 100 \% big burritos . 100 \% . i will definitely be back . great too too is ultra \textsc{<EOT>} &  \\
        \hline 
        \hline 
        \rowcolor{Gray}
(D) & \multicolumn{2}{l}{\textbf{(5 stars)} \textsc{<BOT>} delicious ! the sandwiches are really good and the meat is top quality . it ' s also nice grabbing an exotic item from the shelf for dessert .}  \\
        \hline 
(M) & delicious ! the meat really are good and the quality is nice . it ' s also tempting top notch lovers from the roasters an item top . \textsc{<EOT>} & delicious ! the sandwiches are really good and the quality is top notch . it ' s an exotic item popping also generates from the top spices . \textsc{<EOT>} \\
\hdashline[1pt/1pt]
(R) & delicious ! the sandwiches are really good and the meat is quality . it ' s also nice dessert for shipping from the top floor an unhygienic machine . \textsc{<EOT>} & delicious ! the sandwiches are really good and the quality is top notch . it ' s also charging an item assortment from the grocery store for dessert . \textsc{<EOT>} \\
\hdashline[1pt/1pt]
(P) & delicious sandwiches ! the servers are really good and the quality is top notch . it ' s also an item for meat quality memories . \textsc{<EOT>} & who ! the meat are really good and the quality is top notch ' s . it also seems top notch item has yet and an unexpected range for the pistachio . i do cross like john tomatoes from my experience . \textsc{<EOT>} \\
\hdashline[1pt/1pt]
(AE) & delicious ! the sandwiches are really good and the quality is top notch . it ' s also caught meat also fixing an item from the top for nice hash . \textsc{<EOT>} &  \\
    \end{tabular}
    }
    \caption{
    Reconstruction results for models trained on \textbf{Yelp15}.
    We denote:
    (D) Data sample; 
    (M) Mean (latent) reconstruction;
    (R) Reconstruction;
    (P) Perturbed (latent) reconstruction;
    (AE) Reconstruction of AE.
    }
    \label{tab:nlp-recon-yelp-1}
\end{table}

\begin{table}[th]
    \centering
    \setlength{\tabcolsep}{0.5em} 
    {\scriptsize
    \renewcommand{\arraystretch}{1.2}
        \begin{tabular}{l| p{8cm}|p{8cm} }
        \hline \hline
        & sMIM (1024) & sMIM (1024) \textsuperscript{\textdagger} \\
        \hline \hline
        \rowcolor{Gray}
(D) & \multicolumn{2}{l}{\textbf{(Sports)} \textsc{<BOT>} are you regular or goofy ? regularly goofy}  \\
        \hline 
\hdashline[1pt/1pt]
(M) & are you regular or regular ? regular \textsc{<EOT>} & are you regular or regularly ? regular johnny \textsc{<EOT>} \\
\hdashline[1pt/1pt]
(R) & are you regular regular or nintendo ? regular icecream \textsc{<EOT>} & are you regular or regularly ? regularly gethsemane \textsc{<EOT>} \\
\hdashline[1pt/1pt]
(P) & are you or regular worms regular ? regular goldfish by benjamin \textsc{<EOT>} & are you regular or early regularly regularly regularly \textsc{<EOT>} \\
\hdashline[1pt/1pt]
(AE) & are you sex or two frustrated \textsc{<EOT>} &  \\
        \hline 
        \hline 
        \rowcolor{Gray}
(D) & \multicolumn{2}{l}{\textbf{(Health)} \textsc{<BOT>} how do you start to like yourself ? i was taught by my parents .}  \\
        \hline 
\hdashline[1pt/1pt]
(M) & how do you start to like yourself ? i would like to meet my parents by . \textsc{<EOT>} & how do you start to like yourself ? i was taught by my parents . \textsc{<EOT>} \\
\hdashline[1pt/1pt]
(R) & how do you start to yourself like ? i was taught my parents by parents . \textsc{<EOT>} & how do you start to like yourself ? i was taught by my parents . \textsc{<EOT>} \\
\hdashline[1pt/1pt]
(P) & how do you start to like yourself ? i am 27 by my self . \textsc{<EOT>} & how do you start to like yourself ? start by i was taught my foot . \textsc{<EOT>} \\
\hdashline[1pt/1pt]
(AE) & how do you like to after by christmas day ? i like to aid my boss by my brother and state ! \textsc{<EOT>} &  \\
        \hline 
        \hline 
        \rowcolor{Gray}
(D) & \multicolumn{2}{l}{\textbf{(Business \& Finance)} \textsc{<BOT>} how can i find someone in spain ? i'm in spain today , what do you want ?}  \\
        \hline 
\hdashline[1pt/1pt]
(M) & how can i find someone in spain ? i'm in harlem limo , now what do you want ? \textsc{<EOT>} & how can i find someone in spain ? spain in spain ? i'm talking , what did you want ? \textsc{<EOT>} \\
\hdashline[1pt/1pt]
(R) & where can i find someone in spain ? in spain today , what do you want ? \textsc{<EOT>} & how can i find someone in spain ? spain in spain today , what do you want ? \textsc{<EOT>} \\
\hdashline[1pt/1pt]
(P) & how can i find someone in stone ? in nassau i'm sure civilian , what ? you want today ! \textsc{<EOT>} & how can i find someone in spain ? i'm in spain today ? what maytag , do you think ? \textsc{<EOT>} \\
\hdashline[1pt/1pt]
(AE) & how can i find someone in africa investment , ca ? working 6.0 in future with susan toughie \textsc{<EOT>} &  \\
    \end{tabular}
    }
    \caption{
    Reconstruction results for models trained on \textbf{Yahoo Answers}.
    We denote:
    (D) Data sample; 
    (M) Mean (latent) reconstruction;
    (R) Reconstruction;
    (P) Perturbed (latent) reconstruction;
    (AE) Reconstruction of AE.
    }
    \label{tab:nlp-recon-yahoo-1}
\end{table}

Here we provide reconstruction results for PTB (Fig.\ \ref{tab:nlp-recon-ptb-1}),
Yelp15 (Fig.\ \ref{tab:nlp-recon-yelp-1}), and Yahoo Answers (Fig.\ \ref{tab:nlp-recon-yahoo-1}).
Each figure shows (D) Data sample; 
(M) Mean (latent) reconstruction (\ie, $\z_i = \E{}{\Menc(\z|\x_i)}$);
(R) Reconstruction (\ie, $\z_i \sim \Menc(\z|\x_i)$);
(P) Perturbed (latent) reconstruction (\ie, $\z_i \sim \Menc(\z|\x_i;\mu_i, 10\sigma_i)$);
(AE) Reconstruction of AE.
We compare the best performing sMIM model to an AE with the same architecture, and to sMIM (1024) \textsuperscript{\textdagger}
(\ie, the model trained on the Everything dataset). 

Interestingly, AEs tend to perform worse for longer sentences, when compared to sMIM.
We attribute this to the higher latent entropy, which leads to non-semantic errors 
(\ie, nearby latent codes are less similar compared to MIM).
Another interesting point is how the reconstruction (R), is better in many cases than 
the reconstruction given the mean latent code from the encoder (M) (\ie, which have the highest probability density).
We attribute that to the fact that most probability mass in a high dimensional Gaussian in $d >> 1$ dimensional space and $\sigma$ standard deviation is concentrated in around a sphere of radius $r \approx \sigma \sqrt{d}$. 
As a result the probability mass around the mean is low, and sampling from the mean is less likely to represent the input sentence $\x_i$.
This also explains how perturbations of up to 10 standard deviations might result in good reconstructions.
Finally, we point how sMIM (1024) \textsuperscript{\textdagger}, trained on Everything, does a better job handling longer sentences.

\FloatBarrier

\subsection{Interpolation}

\begin{table}[th]
    \centering
    \setlength{\tabcolsep}{0.5em} 
    {\scriptsize
    \renewcommand{\arraystretch}{1.2}
        \begin{tabular}{p{8.2cm}|p{8.2cm}}
        \hline \hline
        sMIM (512) & sMIM (1024) \textsuperscript{\textdagger} \\
        \hline \hline
        \rowcolor{Gray}
\multicolumn{2}{l}{\textsc{<BOT>} thanks to modern medicine more couples are growing old together} \\
\hline
\textbullet ~ to growing small businesses are growing more rapidly growing \textsc{<EOT>} & \textbullet ~ thanks to modern medicine more modern couples are growing together than \textsc{<EOT>}\\
\textbullet ~ growing to more areas are growing preventing black trends \textsc{<EOT>} & \textbullet ~ thanks to modern cancer more are growing peaceful couples form \textsc{<EOT>}\\
\textbullet ~ growing to the growing industry are growing more rapidly growing than \textsc{<EOT>} & \textbullet ~ thanks to medicine rosen modern more are growing together governing \textsc{<EOT>}\\
\textbullet ~ growing to the exact industry has been growing more sophisticated six months \textsc{<EOT>} & \textbullet ~ thanks to moolah the modern premises are more sensitive together \textsc{<EOT>}\\
\textbullet ~ politics the growing issue are not to mention closely although other prospective products \textsc{<EOT>} & \textbullet ~ programm thanks to the cutbacks schedules is not an church system \textsc{<EOT>}\\
\textbullet ~ the system is growing enough to make not radical an article \textsc{<EOT>} & \textbullet ~ humana remains the loyalty to instituting dynamic is an orthodox montage \textsc{<EOT>}\\
\textbullet ~ the system is reducing compliance not to consider an article \textsc{<EOT>} & \textbullet ~ the strategies is not paying the non-food system an individual member \textsc{<EOT>}\\
\textbullet ~ the system is the problem system not an effective \textsc{<EOT>} & \textbullet ~ the system is not the individual problem member an can \textsc{<EOT>}\\
\textbullet ~ the system is the system not knowing an individual \textsc{<EOT>} & \textbullet ~ the system is not the individual problem an individual member \textsc{<EOT>}\\
\textbullet ~ the system is the system not an encouraging problem \textsc{<EOT>} & \textbullet ~ the system is not the individual problem an individual member \textsc{<EOT>}\\
\hline
        \rowcolor{Gray}
 \multicolumn{2}{l}{\textsc{<BOT>} the system is the problem not an individual member}  \\
\hline
\textbullet ~ the system is the system not an investment fund \textsc{<EOT>} & \textbullet ~ the system is the ringers not an individual member \textsc{<EOT>}\\
\textbullet ~ the system is the problem not an office \textsc{<EOT>} & \textbullet ~ the system is not the problem an individual member \textsc{<EOT>}\\
\textbullet ~ the system is not the problem for an individual \textsc{<EOT>} & \textbullet ~ the problem is not the indies system an individual \textsc{<EOT>}\\
\textbullet ~ the system is not clear the veto \textsc{<EOT>} & \textbullet ~ the merksamer is not the problem system an individual \textsc{<EOT>}\\
\textbullet ~ the system is not encouraging to the securities \textsc{<EOT>} & \textbullet ~ mr . the herald is not an individual problem \textsc{<EOT>}\\
\textbullet ~ xtra the system is not even critical \textsc{<EOT>} & \textbullet ~ qintex producers is the president's to comment \textsc{<EOT>}\\
\textbullet ~ sony denies the declines to secure \textsc{<EOT>} & \textbullet ~ sony preferences itself is the bidding to comment \textsc{<EOT>}\\
\textbullet ~ everyone brought the stock to comment \textsc{<EOT>} & \textbullet ~ sony sony itself is to comment \textsc{<EOT>}\\
\textbullet ~ sony which declines to comment \textsc{<EOT>} & \textbullet ~ sony sony itself to comment \textsc{<EOT>}\\
\textbullet ~ kellogg declines to induce itself \textsc{<EOT>} & \textbullet ~ sony declines itself to sony \textsc{<EOT>}\\
\hline
        \rowcolor{Gray}
 \multicolumn{2}{l}{\textsc{<BOT>} sony itself declines to comment} 
 \\ \hline \hline
\end{tabular}
    }
    \caption{
    Interpolation results between latent codes of input sentences (with gray) from \textbf{PTB}.
    }
    \label{tab:nlp-interp-ptb-1}
\end{table}

\begin{table}[th]
    \centering
    \setlength{\tabcolsep}{0.5em} 
    {\scriptsize
    \renewcommand{\arraystretch}{1.2}
        \begin{tabular}{p{8.2cm}|p{8.2cm}}
        \hline \hline
        sMIM (1024) & sMIM (1024) \textsuperscript{\textdagger} \\
        \hline \hline
        \rowcolor{Gray}
\multicolumn{2}{l}{\textbf{(3 star)} \textsc{<BOT>} as bbq in phoenix goes - this is one of the better ones . get there early - they fill up fast !} \\
\hline
\textbullet ~ as in china phoenix - this is one of the better ones fast get . fill there early - they fill up early ! \textsc{<EOT>} & \textbullet ~ as in phoenix goes this is - better than one of the newest ones . get there early - they fill up fast ! \textsc{<EOT>}\\
\textbullet ~ as far in san jose - this is one of the better ones . fast get up early ! there they fill up fast for u ! \textsc{<EOT>} & \textbullet ~ as shore goes in phoenix - this is one of the better bbq . fast ! they get up there early - men dinner . \textsc{<EOT>}\\
\textbullet ~ as pei wei goes in this phoenix - - one of the best ones . get there early ! they picked up fast food items is better . \textsc{<EOT>} & \textbullet ~ as dean goes in phoenix this is the list of bbq . - one not goes fast - get there early ! they fill up fast . \textsc{<EOT>}\\
\textbullet ~ oxtail yo buffet in pittsburgh as the owners goes - better . this is not one of those fast food places . fill up there get the hot ! \textsc{<EOT>} & \textbullet ~ veal as rocks as this goes in the phoenix area . - one of food is not better quick enough they get . 2 enchiladas up ! \textsc{<EOT>}\\
\textbullet ~ ah circle k ! not as bad in the food . thankfully - this one is one of the best bbq joints here ! service was fast friendly . \textsc{<EOT>} & \textbullet ~ kohrs as molasses as comparing goes in the food . not sure is one of this better ones - the only ones for fat . thumbs squeeze there ! \textsc{<EOT>}\\
\textbullet ~ ehh = ciders as the food goes . not bad for service ! - in many fast the only ones available is this . you can get better steak anywhere else ! \textsc{<EOT>} & \textbullet ~ omg = rainbow not as the food goes . congrats service ! this is one of the hot spots for only frozen hot - you can . eat on carts there . \textsc{<EOT>}\\
\textbullet ~ bin spaetzle food not the best . wicked spoon ! service is brutal only fast for the hot mexican in lv . everything else on this planet as can you get . \textsc{<EOT>} & \textbullet ~ = frozen food ! not the best . only frozen hot as for you shall pick the ice cream - . loved everything else on wednesday ! \textsc{<EOT>}\\
\textbullet ~ frankie food not soo the best . service = horrible ! only drawback frozen for these hike . everything you can pass on the juke planet . \textsc{<EOT>} & \textbullet ~ = food not only the best . frozen service ! everything else for the frozen yogurt company . absolute hot tea during normal on as they can . \textsc{<EOT>}\\
\textbullet ~ food not the best service . knocking only 99 cents ! for the hot buffet everything . beef \& broccoli on the vip polo you can pass . \textsc{<EOT>} & \textbullet ~ = food not . the best frozen service ! only five stars for the water suppose . hot things you can smell on budget . \textsc{<EOT>}\\
\textbullet ~ food not the best . service = horrible ! only plopped for the paella everything \& rum . you can find everything on the strip . \textsc{<EOT>} & \textbullet ~ food = not the best . frozen service ! only \$ 21 for the frozen hot chocolate . everything else can you tell on romance . \textsc{<EOT>}\\
\hline
        \rowcolor{Gray}
 \multicolumn{2}{l}{\textbf{(2 star)} \textsc{<BOT>} food = not the best . service = horrible ! only known for the frozen hot chocolate . everything else you can pass on .}  \\
\hline
\textbullet ~ food not the best . fuck service only ! ! horrible cannolis for the fajitas unusual known . everything you can pass on graduate . \textsc{<EOT>} & \textbullet ~ food = not the best . frozen hot service ! only website for the frozen hot chocolate . you can grab everything else on . \textsc{<EOT>}\\
\textbullet ~ food not suck . the best service ever ! just horrible everything for the frozen hot chocolate . you can probably survive on everything else . \textsc{<EOT>} & \textbullet ~ food = not the best . frozen service ! only for five stars during the san francisco frozen chicken . everything else on could not give thumbs . \textsc{<EOT>}\\
\textbullet ~ food = not ! service = the best . only organizations thing for chocolate lovers treats and green beans . everything you can taste on the planet . \textsc{<EOT>} & \textbullet ~ food = not ! the frozen yogurt . service only best for you ate here twice although the frozen yogurt . delicious atmosphere on everything else . \textsc{<EOT>}\\
\textbullet ~ blech food ! not the best dish anywhere else . service = <unk> for the frozen hot chocolate and dessert bartenders ! everything you can only expect better at this shuffle . \textsc{<EOT>} & \textbullet ~ gelato food ! not sure the best . frozen seared only wish you can mix for the frozen hot chocolate frozen . service on and everything else explains . \textsc{<EOT>}\\
\textbullet ~ 32 words ! not amazing food . the best <unk> music and service they had can earned a better meal at xs . everything else on bill for me . \textsc{<EOT>} & \textbullet ~ hilariously = ! food is not the best meal . hibachi cover service and they only wished a frozen yogurt for hot girl . better luck at <unk> and on the latter experience . \textsc{<EOT>}\\
\textbullet ~ snottsdale act ! ! rio mia <unk> at the food and wished you not a fan . delicious lunch \& dessert better choices for dessert but they had blackjack . \textsc{<EOT>} & \textbullet ~ blended ! wifey better food ! the service is not frozen hot . they redeemed a <unk> and only frozen someplace at horse's for frozen worms . \textsc{<EOT>}\\
\textbullet ~ husbands cher ! wish they had <unk> dessert at the bellagio and not a great lunch selection . food better tasting wise but sadly serves and dessert selection . \textsc{<EOT>} & \textbullet ~ wish ! methinks buffet is ingrediants at the <unk> food and a better tasting . they woulda frozen lunch but not memorable and satisfying tasting better ambiance . \textsc{<EOT>}\\
\textbullet ~ soooo ! pretzel panera <unk> they had at a better selection and the food sucked but nothing memorable a dessert . surely great value and better mayonnaise desserts . \textsc{<EOT>} & \textbullet ~ yummy ! wish they had <unk> at a buffet and netherlandish better tasting food . a renovation treasure and great value but not better than calories tasting . \textsc{<EOT>}\\
\textbullet ~ yummy ! wish they had <unk> at lunch and a dessert selection but a better value and great value than beef suggestion company . \textsc{<EOT>} & \textbullet ~ wish ! wish they had <unk> at 10am and a dessert selection but better food a better and better tasting selection . great value ! \textsc{<EOT>}\\
\textbullet ~ yummy ! wish they had <unk> dessert at lunch and a selection but a tiramisu better value and freshness value food taste better than ihop . \textsc{<EOT>} & \textbullet ~ wish ! wish they had lunch at <unk> and a dessert fountain but better than a selection and great tasting food servings better tasting . \textsc{<EOT>}\\
\hline
        \rowcolor{Gray}
 \multicolumn{2}{l}{\textbf{(4 star)} \textsc{<BOT>} yummy ! wish they had <unk> at lunch and a better dessert selection but a great value and better tasting food than wicked spoon .} 
 \\ \hline \hline
\end{tabular}
    }
    \caption{
    Interpolation results between latent codes of input sentences (with gray) from \textbf{Yelp15}.
    }
    \label{tab:nlp-interp-yelp-1}
\end{table}

\begin{table}[th]
    \centering
    \setlength{\tabcolsep}{0.5em} 
    {\scriptsize
    \renewcommand{\arraystretch}{1.2}
        \begin{tabular}{p{8.2cm}|p{8.2cm}}
        \hline \hline
        sMIM (1024) & sMIM (1024) \textsuperscript{\textdagger} \\
        \hline \hline
        \rowcolor{Gray}
\multicolumn{2}{l}{\textbf{(Business \& Finance)} \textsc{<BOT>} are u shy or outgoing ? both , actually} \\
\hline
\textbullet ~ are u or wishing vidio ? both , actually \textsc{<EOT>} & \textbullet ~ are u shy or k ? both , actually \textsc{<EOT>}\\
\textbullet ~ are u or stressed caffiene ? both , actually make a smile \textsc{<EOT>} & \textbullet ~ are u minded or rem ? actually , both \textsc{<EOT>}\\
\textbullet ~ witch are u or how lucky ? both \textsc{<EOT>} & \textbullet ~ are u transparent or shy ? it'd actually , add-on \textsc{<EOT>}\\
\textbullet ~ are u kidding or spraying ? both \textsc{<EOT>} & \textbullet ~ are u untouchable cubed or programe ? both , actually like \textsc{<EOT>}\\
\textbullet ~ how does wile or are you ? to both use , instead like it . \textsc{<EOT>} & \textbullet ~ wha do u are roselle or marketed ? you start , by both my inbox \textsc{<EOT>}\\
\textbullet ~ how do u choose to start or ? like i cant think , are actually better by my work . \textsc{<EOT>} & \textbullet ~ how do u simplify phases towards you ? are proving , like no smiles . \textsc{<EOT>}\\
\textbullet ~ how do you start to alienate yourself ? i are like or drone , my actually feels . \textsc{<EOT>} & \textbullet ~ how do you burp confidence ? to start i was like , shareaza the new by hindering . \textsc{<EOT>}\\
\textbullet ~ how do you start to yourself or like ? i like my math side . \textsc{<EOT>} & \textbullet ~ how do you start to race ? i like kazaa when my was cheated . \textsc{<EOT>}\\
\textbullet ~ how do you start to like yourself ? i think my parents is by focusing . \textsc{<EOT>} & \textbullet ~ how do you start to start like ? i was taught by my parents . \textsc{<EOT>}\\
\textbullet ~ how do you start to yourself like ? i was taught by my parents . \textsc{<EOT>} & \textbullet ~ how do you start to like yourself ? i was taught by my parents . \textsc{<EOT>}\\
\hline
        \rowcolor{Gray}
 \multicolumn{2}{l}{\textbf{(Health)} \textsc{<BOT>} how do you start to like yourself ? i was taught by my parents .}  \\
\hline
\textbullet ~ how do you start to yourself by allowing ? i like my parents yr . \textsc{<EOT>} & \textbullet ~ how do you start to like yourself ? i was taught by new england . \textsc{<EOT>}\\
\textbullet ~ how do you start to yourself like i ? my parents was by mario practitioner . \textsc{<EOT>} & \textbullet ~ how do you start to like yourself ? i was taught by my parents . \textsc{<EOT>}\\
\textbullet ~ how do you start to cite yourself ? i like by my consequences in 1981 . \textsc{<EOT>} & \textbullet ~ how do i start you to beethoven ? like israel was my grandmother by fielders . \textsc{<EOT>}\\
\textbullet ~ how do i start girls like to ? you can find yourself in my states , by today . \textsc{<EOT>} & \textbullet ~ how do you start to find ? i like aggieland in my testicles was listening . \textsc{<EOT>}\\
\textbullet ~ how do you start yourself drunk ? i can find in something like to my country , what by jane . \textsc{<EOT>} & \textbullet ~ how can i do compuserve attain ? start to comment in spain you like , was my real pics . \textsc{<EOT>}\\
\textbullet ~ how can i start those neeed in america ? do you like to rephrase an invention , what i'm spinning ? \textsc{<EOT>} & \textbullet ~ how can i find blueprints do you ? i'm in spain like queens to chelsea , arrange . \textsc{<EOT>}\\
\textbullet ~ how can i find someone in spain ? i'm guessing today by pascal , what do you want to ? \textsc{<EOT>} & \textbullet ~ how can i find uneasy profiles in spain ? i'm sure what you do , like today's ? \textsc{<EOT>}\\
\textbullet ~ how can i find an attorney in spain ? i'm studying chicken's what , do you want to ? \textsc{<EOT>} & \textbullet ~ how can i find someone in spain ? i'm in spain today , what do you want ? \textsc{<EOT>}\\
\textbullet ~ how can i find someone in spain ? in spain i'm studying , what do you want ? \textsc{<EOT>} & \textbullet ~ how can i find someone in spain ? i'm in tanks today , what do you want to ? \textsc{<EOT>}\\
\textbullet ~ how can i find someone in spain ? i'm in italy today , what do you want ? \textsc{<EOT>} & \textbullet ~ how can i find someone in spain ? i'm guessing in spain today , what do you want ? \textsc{<EOT>}\\
\hline
        \rowcolor{Gray}
 \multicolumn{2}{l}{\textbf{(Business \& Finance)} \textsc{<BOT>} how can i find someone in spain ? i'm in spain today , what do you want ?} 
 \\ \hline \hline
\end{tabular}
    }
    \caption{
    Interpolation results between latent codes of input sentences (with gray) from \textbf{Yahoo Answers}.
    }
    \label{tab:nlp-interp-yahoo-1}
\end{table}

Here we provide interpolation results for PTB (Fig.\ \ref{tab:nlp-interp-ptb-1}),
Yelp15 (Fig.\ \ref{tab:nlp-interp-yelp-1}), and Yahoo Answers (Fig.\ \ref{tab:nlp-interp-yahoo-1}).
We compare the best performing sMIM model to sMIM (1024) \textsuperscript{\textdagger}.
Interestingly, both models appear to have learned a dense latent space, 
with sMIM (1024) \textsuperscript{\textdagger} roughly staying within 
the domain of each dataset. This is surprising since the latent space of sMIM (1024) \textsuperscript{\textdagger}
jointly represents all datasets.

\FloatBarrier

\subsection{Sampling}


\begin{table}[th]
    \centering
    \setlength{\tabcolsep}{0.5em} 
    {\scriptsize
    \renewcommand{\arraystretch}{1.2}
        \begin{tabular}{p{16cm}}
        \hline \hline
        sMIM (512)  \\
        \hline \hline
\textbullet ~ instead the stock market is still being felt to <unk> those of our empty than in a bid \textsc{<EOT>} \\
\textbullet ~ he estimated the story will take <unk> of paper co . ' s \$ n million in cash and social affairs to at the company a good share \textsc{<EOT>} \\
\textbullet ~ long-term companies while the company ' s <unk> provisions would meet there to n or n cents a share and some of costly fund \textsc{<EOT>} \\
\textbullet ~ time stocks the company explained him to sell <unk> properties of high-grade claims which has received a net loss in the firm \textsc{<EOT>} \\
\textbullet ~ what i had the recent competition of <unk> replies that is n't expected to draw a very big rise in tokyo \textsc{<EOT>} \\
\end{tabular}
    }
    \caption{
    Samples from best performing model for dataset \textbf{PTB}.
    }
    \label{tab:nlp-sample-ptb-1}
\end{table}

\begin{table}[th]
    \centering
    \setlength{\tabcolsep}{0.5em} 
    {\scriptsize
    \renewcommand{\arraystretch}{1.2}
        \begin{tabular}{p{16cm}}
        \hline \hline
        sMIM (1024)  \\
        \hline \hline
\textbullet ~ ben monkey gabi sister near the western fest . i ' ve been looking forward to this location , and each time i ' m in the 6th bunch i want to have a great visit experience . it was all kinds of fillers , owns and dressings non-asian with jalapeños <unk> does n't hold me for much healthier . front desk is not my favorite dinner place at the gates . they are closed on mondays , - lrb - it could affect a couple minutes more rocks - rrb - and then we said the bar was the real bold . i ' d rather go to firefly some bubble in greece . if you had a neighbourhood addiction <unk> c , take this look as most amazing . \textsc{<EOT>} \\
\textbullet ~ hello tanya stephen covering qualité . ugh haha , i was curious to consume that the white asian restaurants believes filled a mob and turkey melt departments for \$ 9.99 . the <unk> of these were not intrusive , it was accepted in there . . i ' m sure this is n't one of my favorite places to go at night with here ! particularly speaking the italian cleaning tables . we also ordered some pina colada , which tasted exactly like they came out of a box and per endearing thick . pretty good food overall , and the pigeons self nightly . i ' d call it again just on halloween for a dependable lunch . but the statue sucks ? so if you have bouchon to inquire was good place . \textsc{<EOT>} \\
\textbullet ~ prada based pata based solely often inside . this place is unappealing horrific for the 50th and fries , i ' ve caught to have a ton of good reviews <unk> in buckeye , barnes knew . not bc that i was wrong with my team being kicked the whole thing at eggroll , it ' s like pulling out of the landmark . no luck on ketchup top crunch , if you are craving something simple and <unk> . we also tried the wild mushroom - lrb - it ' s burn , did n't go in disheveled - rrb - as a matter destination from flavor . the food was just ok and nothing to write home about . friend peeps i only had one beer , but this place does not deserve the same increase . \textsc{<EOT>} \\
\end{tabular}
    }
    \caption{
    Samples from best performing model for dataset \textbf{Yelp15}.
    }
    \label{tab:nlp-sample-yelp-1}
\end{table}

\begin{table}[th]
    \centering
    \setlength{\tabcolsep}{0.5em} 
    {\scriptsize
    \renewcommand{\arraystretch}{1.2}
        \begin{tabular}{p{16cm}}
        \hline \hline
        sMIM (1024)  \\
        \hline \hline
\textbullet ~ how does transformers send grow ina under pubs ? i found the suspension resides official game is exciting to withstand and what can a person do in that case ? brees fights , if it does 150 . the dre is tied ordered outlook <unk> 2005 . today had a migrane with limitation tops , because of his vr repeats , you are referring to review at the university of 1994 and have visited fortune . judy for websites <unk> website is beware confused . \textsc{<EOT>} \\
\textbullet ~ how do i download jesus gyno to woman whom ? being irvine in line is what you did a lot of oceanic denny in the middle east and spanish wallet or <unk> entity . plus , i'm aware of that , particularly do you have any insight insight ... if you are a hoe who's right click on it , and you can ’ t get some skills god . the other government also happened to be <unk> with most varied life-forms is located at this point . foreigners your covers , and maybe even my friends . \textsc{<EOT>} \\
\textbullet ~ what's mastering marathons fluently is einstein among the waivers ? ok i feel that what happened to tom during the holidays moniter of 1-2 awol whn reservoir <unk> . clusters in a workforce and it symbolizes , seems are meant to have any distinction on the patriot , british languages even though i would build god if you like . just bringing your old door as a distorted spree ? hmmmm , because you're not anti-bacterial pure dino and <unk> this can be deduced . \textsc{<EOT>} \\
\end{tabular}
    }
    \caption{
    Samples from best performing model for dataset \textbf{Yahoo Answers}.
    }
    \label{tab:nlp-sample-yahoo-1}
\end{table}

Here we show samples from the best performing models learned from a single dataset  
for PTB (Fig.\ \ref{tab:nlp-sample-ptb-1}), Yelp15 (Fig.\ \ref{tab:nlp-sample-yelp-1}), and Yahoo Answers (Fig.\ \ref{tab:nlp-sample-yahoo-1}).
We sample from a zero-mean Gaussian distribution over the latent space, 
with an isotropic covariance with a standard deviation of 0.1
(since we cannot directly sample from the implicit marginal over the latent).
Interestingly, this simple heuristic provides good samples. 
We attribution this to the anchor, which defines scale and position for the implicit 
marginal over the latent to roughly match.

\FloatBarrier

\subsection{Question Answering}

\begin{table}[th]
    \centering
    \setlength{\tabcolsep}{0.5em} 
    {\scriptsize
    \renewcommand{\arraystretch}{1.2}
        \begin{tabular}{l | p{16cm}}
        \hline \hline
        \multicolumn{2}{l}{sMIM (1024)}   \\
        \hline \hline
(Q) & \textsc{<BOT>} where can i find the most affordable roses for valentine's day this late in the game ? \\
\hdashline[1pt/1pt]
(A) short & \textbullet ~ try the mall , \textsc{<EOT>} \\
(A) medium & \textbullet ~ the game is to play the game , set your mind \textsc{<EOT>} \\
(A) long & \textbullet ~  i think the best answer is to ask her out , especially since the late 1990s game \textsc{<EOT>} \\
        \hline 
(Q) & \textsc{<BOT>} how do you start to like yourself ? \\
\hdashline[1pt/1pt]
(A) short & \textbullet ~ start by yourself , take the steps \textsc{<EOT>} \\
(A) medium & \textbullet ~ start by yourself . you have to start asking yourself why this is happening \textsc{<EOT>} \\
(A) long & \textbullet ~ start by yourself . you have to start acting yourself in the middle of your eye .  \textsc{<EOT>} \\
        \hline 
(Q) & \textsc{<BOT>} how can i find someone in spain ? \\
\hdashline[1pt/1pt]
(A) short & \textbullet ~ try classmates.com or classmates.com to see \textsc{<EOT>} \\
(A) medium & \textbullet ~ someone who is a good one to talk to your question \textsc{<EOT>} \\
(A) long & \textbullet ~  someone who is a good person to find a nice girl , and take care of yourself \textsc{<EOT>} \\
        \hline 
(Q) & \textsc{<BOT>} with a weapon to your person , who in you family do you save with an answer ? \\
\hdashline[1pt/1pt]
(A) short & \textbullet ~ you should save yourself in person \textsc{<EOT>} \\
(A) medium & \textbullet ~ you can save the money in person , or do you want to save money \textsc{<EOT>} \\
(A) long & \textbullet ~ you can save your money in person , and answer the question yourself . do you have a family ?  \textsc{<EOT>} \\
        \hline 
(Q) & \textsc{<BOT>} what are the best sites to shop for personal computers online ? \\
\hdashline[1pt/1pt]
(A) short & \textbullet ~ try amazon.com online . \textsc{<EOT>} \\
(A) medium & \textbullet ~ i think it is best to shop online , or take a look at the personal \textsc{<EOT>} \\
(A) long & \textbullet ~ yahoo is best online . i would suggest checking out the personal website for personal info  \textsc{<EOT>} \\
        \hline 
(Q) & \textsc{<BOT>} who is going to win the super bowl this year ? \\
\hdashline[1pt/1pt]
(A) short & \textbullet ~ the steelers is a pretty good \textsc{<EOT>} \\
(A) medium & \textbullet ~ the pittsburgh steelers is a good one , but i don't know \textsc{<EOT>} \\
(A) long & \textbullet ~ this is the best team to win the super bowl , and i think you mean good luck  \textsc{<EOT>} \\
        \hline 
(Q) & \textsc{<BOT>} is there a web site that provides info on companies that have been known to provide lousy service ? \\
\hdashline[1pt/1pt]
(A) short & \textbullet ~ yes , google was a little service \textsc{<EOT>} \\
(A) medium & \textbullet ~ i have known as a service that provides a service pack to provide transparency . \textsc{<EOT>} \\
(A) long & \textbullet ~ try searching on google and search for that info . there are many different types of service that provide to the service that has been answered  \textsc{<EOT>}  \\
        \hline 
(Q) & \textsc{<BOT>} what is the closest capital city to paris ? \\
\hdashline[1pt/1pt]
(A) short & \textbullet ~ paris hilton city , ca \textsc{<EOT>} \\
(A) medium & \textbullet ~ in the south city , ca , i think you mean the city hall \textsc{<EOT>} \\
(A) long & \textbullet ~ city is the capital city in south america . i think it is a beautiful city  \textsc{<EOT>} \\
        \hline 
(Q) & \textsc{<BOT>} is santa claus real ? \\
\hdashline[1pt/1pt]
(A) short & \textbullet ~ yes , you can ask her to questions \textsc{<EOT>} \\
(A) medium & \textbullet ~ yes , it is a santa claus . the phrase is to take care \textsc{<EOT>} \\
(A) long & \textbullet ~  yes , it is real . santa claus are the best in santa claus . take care \textsc{<EOT>}  \\
        \hline 
(Q) & \textsc{<BOT>} do height increasing devices shown on television works ? \\
\hdashline[1pt/1pt]
(A) short & \textbullet ~ perhaps shown is shown to see \textsc{<EOT>} \\
(A) medium & \textbullet ~ yes , the best way to increase it is to work \textsc{<EOT>} \\
(A) long & \textbullet ~ yes , shown shown . the biggest problem with television is to take a walk away from the edge \textsc{<EOT>} \\
        \hline 
(Q) & \textsc{<BOT>} where can i find advice on hip hop beat making ? \\
\hdashline[1pt/1pt]
(A) short & \textbullet ~ try this website : http://messenger.yahoo.com/ \textsc{<EOT>} \\
(A) medium & \textbullet ~ try this website : http://messenger.yahoo.com/ to find the best answer . \textsc{<EOT>} \\
(A) long & \textbullet ~ try making a video club , such as the lion and hip hop . i wish you to be good  \textsc{<EOT>} \\
        \hline 
(Q) & \textsc{<BOT>} how do u clear the history in the search field ? \\
\hdashline[1pt/1pt]
(A) short & \textbullet ~ clear search history , it will be clear \textsc{<EOT>} \\
(A) medium & \textbullet ~ clear history , it appears that the question period has expired . if you have received an answer that meets your needs , please choose a ' best answer . ' regards \textsc{<EOT>} \\
(A) long & \textbullet ~  clear history in the search field , u can find it from your website : * clear history \textsc{<EOT>} \\
        \hline 
(Q) & \textsc{<BOT>} where can i get chapter summaries in text books ? \\
\hdashline[1pt/1pt]
(A) short & \textbullet ~ try this site , you'll find it interesting \textsc{<EOT>} \\
(A) medium & \textbullet ~ text books ? try this site , and get a book to read \textsc{<EOT>} \\
(A) long & \textbullet ~  in books , it is a text book , and the text books are written in the same text . \textsc{<EOT>} \\
        \hline 
(Q) & \textsc{<BOT>} how to tell a nice guy you dont like him ? \\
\hdashline[1pt/1pt]
(A) short & \textbullet ~ nice guy dont know what to do \textsc{<EOT>} \\
(A) medium & \textbullet ~ nice guy , dont tell him what the hell is \textsc{<EOT>} \\
(A) long & \textbullet ~ dont tell him that you like him ? nice guy , and the guy who is nice to him !  \textsc{<EOT>} \\
        \hline 
(Q) & \textsc{<BOT>} does your body feel physically fit ? \\
\hdashline[1pt/1pt]
(A) short & \textbullet ~ no , it is a little bit \textsc{<EOT>} \\
(A) medium & \textbullet ~ feel your body needs to fit into the body . i feel like a good fit \textsc{<EOT>} \\
(A) long & \textbullet ~  feel your body fit in a fit body . i feel like the best fit to fit in your body \textsc{<EOT>} \\
        \hline 
\end{tabular}
    }
    \caption{
    Question and sampled answers from model sMIM (1024) (\ie, trained on Yahoo Answers dataset). 
    We provide short/medium/long sampled answers (A) for each question (Q). 
    }
    \label{tab:nlp-qa-yahoo-1}
\end{table}

Here we provide more examples of answers generated from a model trained
on Yahoo Answers (\ie, sMIM (1024) in Fig.\ \ref{tab:nlp-qa-yahoo-1}).
In particular, the model was trained from data in which 20\% of the encoder input tokens were 
replaced with the <unk> token.
This is a form of self-supervised learning commonly used in language modelling (\eg, \citet{DBLP:journals/corr/BowmanVVDJB15}).
This encourages the model to replace <unk> with other tokens.
We have found this procedure to  significantly improve  the quality of the generated answers.
We provide three generated answers for each question (Q), taken from Yahoo Answers. 
Short/medium/long answers (A) are generated by concatenating 5/10/15 <unk> tokens.
The number of <unk> encodes the length of the expected answer.
We note that, in many cases, only one answer will be a good match to the question, suggesting the model
has preferences towards answers with a question specific length.

\FloatBarrier





\end{document}